\documentclass[letterpaper]{article} % DO NOT CHANGE THIS
\usepackage{aaai24}  % DO NOT CHANGE THIS
\usepackage{times}  % DO NOT CHANGE THIS
\usepackage{helvet}  % DO NOT CHANGE THIS
\usepackage{courier}  % DO NOT CHANGE THIS
\usepackage[hyphens]{url}  % DO NOT CHANGE THIS
\usepackage{graphicx} % DO NOT CHANGE THIS
\usepackage{natbib}  % DO NOT CHANGE THIS AND DO NOT ADD ANY OPTIONS TO IT
\usepackage{caption} % DO NOT CHANGE THIS AND DO NOT ADD ANY OPTIONS TO IT
\usepackage{algorithm}
\usepackage{algorithmic}

\usepackage{amssymb}
\usepackage{amsmath}
\usepackage{multirow}
\usepackage{booktabs}
\usepackage{subfig}

\usepackage{newfloat}
\usepackage{listings}
\DeclareCaptionStyle{ruled}{labelfont=normalfont,labelsep=colon,strut=off} % DO NOT CHANGE THIS
\floatstyle{ruled}
\newfloat{listing}{tb}{lst}{}
\floatname{listing}{Listing}
\title{Beyond the Label Itself: Latent Labels Enhance Semi-supervised Point Cloud Panoptic Segmentation}
\author{
    Yujun Chen\textsuperscript{\rm 1,\rm 2,\equalcontrib},
    Xin Tan\textsuperscript{\rm 1,\rm 2,\equalcontrib},
    Zhizhong Zhang\textsuperscript{\rm 1,\rm 2,\thanks{Corresponding author.}}, 
    Yanyun Qu\textsuperscript{\rm 3},
    Yuan Xie\textsuperscript{\rm 1,\rm 2}
}
\affiliations{
    \textsuperscript{\rm 1}School of Computer Science and Technology, East China Normal University\\
    \textsuperscript{\rm 2}Chongqing Institute, East China Normal University\\
    \textsuperscript{\rm 3}School of Information Science and Engineering, Xiamen University\\
    51215901065@stu.ecnu.edu.cn, \{xtan, zzzhang, yxie\}@cs.ecnu.edu.cn, 
    yyqu@xmu.edu.cn
}

\usepackage{bibentry}
\begin{document}

\maketitle

\begin{abstract}
As the exorbitant expense of labeling autopilot datasets and the growing trend of utilizing unlabeled data, semi-supervised segmentation on point clouds becomes increasingly imperative. Intuitively, finding out more ``unspoken words'' (i.e., latent instance information) beyond the label itself should be helpful to improve performance.
In this paper, we discover two types of latent labels behind the displayed label embedded in LiDAR and image data.
First, in the LiDAR Branch, we propose a novel augmentation, Cylinder-Mix, which is able to augment more yet reliable samples for training.
Second, in the Image Branch, we propose the Instance Position-scale Learning (IPSL) Module to learn and fuse the information of instance position and scale, which is from a 2D pre-trained detector and a type of latent label obtained from 3D to 2D projection.
Finally, the two latent labels are embedded into the multi-modal panoptic segmentation network.
The ablation of the IPSL module demonstrates its robust adaptability, and the experiments evaluated on SemanticKITTI and nuScenes demonstrate that our model outperforms the state-of-the-art method, LaserMix.
\end{abstract}
% However, the current methods have not deeply explored more profound label information from self-information.

\section{Introduction}

Point clouds, with richer visual and geometric information, have played an increasingly significant role in perception tasks \cite{roriz2021automotive-survey,triess2021survey-lidar}. Given its extensive potential, point cloud panoptic segmentation, unifying instance and semantic segmentation, has been applied in various fields, such as autonomous driving, robotics, and industrial manufacturing \cite{geiger2012benchmark,fernandes2021point-det-survey,wang2019applications}.  
% \cite{choy20194d}

However, it is exhausting to annotate point cloud data, which prohibitively restricts its potential applications \cite{hu2022sqn,unal2022scribble,xu2020weakly-10x}. Hence, it is highly demanded to utilize less point data to achieve better performance. Nowadays, images play a crucial role in point cloud segmentation since they are cost-effective \cite{cui2021deep,li2021deepi2p}. In addition, 
%Here the imperious demands is to utilize less data to achieve better segmentation performance. Meanwhile, images, which are cost-effective, plays a crucial role in point cloud segmentation. Furthermore, 
with the larger number of unlabeled point cloud data collected, the trend towards semi-supervision is prevalent \cite{sindhwani2005beyond,you2022unsupervised}. Thus, in this paper, we pay attention to semi-supervised multi-modal point cloud (panoptic) segmentation (SMPS).

%------------------------------------------------------------------------
\begin{figure}[t]
\begin{center}
\includegraphics[width=0.92\linewidth]{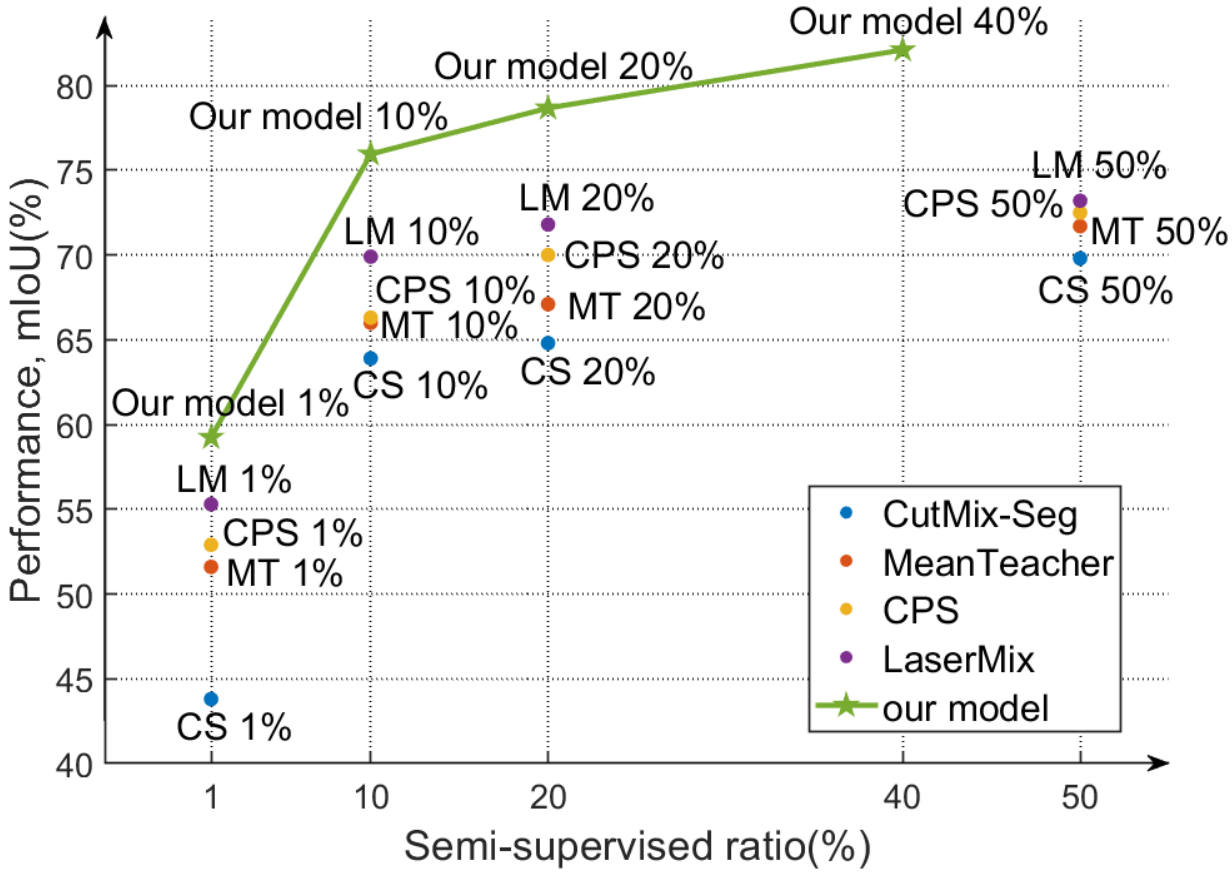}
\end{center}
   \caption{Segmentation quality at various Semi-supervised ratios. Our model outperforms all other methods in mIoU.} 
    % by a large margin
\label{introduction_mIoU}
% \vspace{-1mm}
\end{figure}

%------------------------------------------------------------------------

We have observed two major limitations in the previous SMPS methods. First, due to the limited ground-truth data, the preceding semi-supervised (SS) models generated pseudo-labels as ground-truth for training \cite{kong2023lasermix,park2022DetMatch}. Nevertheless, such pseudo-labels are sometimes unreliable and of poor quality when the network is not fully trained.
% for processing 3D point clouds, network training urgently requires more samples.
Second, for the 3D-2D cross-modal processing, previous multi-modal point cloud models \cite{liu2022bevfusion,liang2022bevfusion-detection} treat all pixels equally without leveraging sufficient image-level information (e.g., size and boundary), despite the fact that images have a superior ability to detect and locate objects, which even outperforms point clouds in certain cases \cite{park2022DetMatch}. 

% it can be concluded that the key challenge lies in extracting information as much as possible within the limited labeled data conditioned on the semi-supervised setting. Luckily, in this paper,
According to the above discussions, we argue that the information contained in point cloud labels is not only as superficial as the displayed label itself but also encompasses some latent labels beneath the surface.
Hence, it is a feasible solution to explore unspoken words (i.e., latent labels) to address the two issues. Specifically, for the first, since only given labeled data is reliable and completely correct, 1) \textit{can we exploit self-information among labeled point clouds to construct other reliable and diverse labels?} 
For the second one, the existing 3D instance labels can be projected onto the image as a form of weak 2D annotations, which indicate the instances' position and scale. Therefore, 2) \textit{can we utilize these weak annotations from 3D-2D self-information to improve the network segmentation performance?}

In this paper, we propose to address the limitations by discovering more latent labels. As illustrated in Figure \ref{introduction_mIoU}, our method significantly surpasses previous semi-supervised point cloud segmentation ones. 

First, we propose a novel data augmentation method, Cylinder-Mix, to obtain reliable and diverse samples for training. Specifically, to exploit self-information within limited data, It performs interleaved mixing for labeled point clouds based on cylinder voxels. This kind of mix strategy does not require other annotations, and the obtained labels are all as accurate as the human-annotated ones. 
% In our experiments, Cylinder-Mix outperforms the augmentation of LaserMix \cite{kong2023lasermix}.

Second, Instance Position-Scale Learning (IPSL) Module is proposed to learn the instance information of position and boundary on images. In this module, latent labels from 3D-2D self-information, the instance boxes, are obtained through LiDAR-to-camera projection. Subsequently, 
inspired by interactive image segmentation (IIS) \cite{ramadan2020IIS-survey,lin2020first-click}, weak interactive annotation (e.g., the bounding box with location and boundary information) leads the network to pay more attention to the highlighted object. Hence, 
the boxes, which serve as interactive annotations, are encoded into heatmaps to indicate positions and scales, and then fused to the 2D backbone. This way, these latent labels from 3D-2D projection highlight the instances so as to improve segmentation.
% Meanwhile, the instance boxes could be predicted after training with LiDAR-to-camera projected boxes,can serve as interactive annotations to improve network segmentation performance. 。 without extra annotation cost. Finally through the encoding and fusion of the predicted weak annotations, the instances information of position and boundary can be fused into the network.}

Furthermore, our experiments demonstrate the adaptability and versatility of the IPSL module, as it still works well whether it fuses instance boxes from a trained detector or masks from a large model (e.g., SAM \cite{kirillov2023segment-anything}, and Grounded-Segment-Anything \cite{GroundedSAM}).
Additionally, to our best knowledge, our model is the first multi-modal model applying the semi-supervised setting to solve point cloud panoptic segmentation.
% can be well-suited to the era of large-scale models since the masks generated from the 3D to 2D projection can be replaced by the large models.

In summary, our contributions are summarized as follows:
\begin{itemize}
    \item We explore latent labels from self-information on both point clouds and images, without requiring any extra labels, to enhance network segmentation performance. Surprisingly, our model finally achieves state-of-the-art performance on semi-supervised semantic segmentation, surpassing the previous best method, LaserMix.
    \item We propose a novel data augmentation technique, Cylinder-Mix, that could obtain reliable and diverse labels, as accurate as the given ground-truth, within limited labeled data for semi-supervised training.
    \item We propose the Instance Position-scale Learning (IPSL) Module to fuse latent annotations from 3D-2D self-information. The obtained latent labels, instance boxes, are utilized to provide position and scale explicitly.
\end{itemize}
%contributions 1. 第一个半监督的全景分割在real-world数据集上的，并且半监督语义分割达到sota 2. 提出了一种新的数据增强方法 3. 提出一种用2D弱标注帮助半监督点云分割的范式，这种弱标注可以零成本获得并能提高实例信息，从而在多模态融合的过程中提升半监督分割性能。

%---------------------------------------------------------------------------

\section{Related Work}

\textbf{Fully-supervised Point Cloud Segmentation.} In recent years, the efficiency of point cloud segmentation has been enhanced mainly by 3D perspective feature extraction \cite{zhou2021panoptic-polarnet,li2022Panoptic-PHNet,fong2022panoptic-benchmark,qi2017pointnet} and multi-modal fusion \cite{liu2022bevfusion,liang2022bevfusion-detection,zhang2023lidar,tan2023positive, liu2021deep}. Cylinder-3D \cite{zhu2021cylinder3d} proposes a novel voxelization approach for the geometric properties of the point cloud, while Panoptic-Polarnet \cite{zhou2021panoptic-polarnet} utilizes the Bird's Eye View (BEV) feature for segmentation and clustering, which not only realizes the panoramic segmentation, but also avoids the occlusion issue between instances. However, the 2D image branch has been somewhat neglected, leaving the network far from better performance during semi-supervised tasks. Although BEVfusion \cite{liu2022bevfusion} combines images, it treats all pixels equally without using instance information. Distinctively, we do not only fuse the image features but also focus more on the instance position and boundary of the image, so as to implicitly enhance the performance of the SS segmentation network in the process of multi-modal fusion.

\textbf{Semi-Supervised Point Cloud Methods.} Several methods for indoor scenes have been proposed. SSPC-Net \cite{cheng2021sspc-net} utilizes pseudo-label propagation through superpixel segmentation, while \citet{jiang2021guided-point-CL} applies Guided Point Contrastive Loss to semi-supervised indoor datasets. However, such methods are unable to effectively handle large-scale real-world datasets, such as SemanticKITTI and nuScenes, due to their large number of points and objects. For methods on real-world datasets, LaserMix \cite{kong2023lasermix}, adapting Mean Teacher \cite{tarvainen2017mean-teacher} architecture to SS point segmentation, proposes a new circular mixture data augmentation. However, the generated samples from LaserMix can be unreliable due to the mixture of unlabeled data since the network is not always well-trained. DetMatch \cite{park2022DetMatch} applies the Mean Teacher architecture to detection tasks, but lacks sufficient adaptability for panoptic segmentation like Panoptic-PolarNet. In contrast, our proposed data augmentation, Cylinder-Mix, not only generates reliable and diverse labels from limited labeled data for training but also inherits the excellent feature extraction capability of Cylinder3D, thus making it adaptable for semi-supervised panoptic segmentation tasks on the real-world dataset.

% Previous researches on point cloud segmentation has utilized classic semi-supervised network structure like Mean Teacher \cite{tarvainen2017mean-teacher} and label consistency constraints via data augmentation\cite{park2022DetMatch,kong2023lasermix}. Specifically, Mean Teacher trains a teacher network and a student network with labeled and unlabeled data, and ensures consistency between the two networks for the unlabeled data. 
 % Additionally, current approaches also pay attention of leveraging self-supervised information to extract features. Point-M2AE\cite{zhang2022point-M2AE} applies MAE architecture on point cloud. But it's only managed satisfactory results on simple datasets like ShapeNet\cite{chang2015shapenet} and unable to cope with real-world datasets, such as SemanticKITTI\cite{geiger2012kitti} and nuScenes\cite{caesar2020nuscenes}, due to their considerable number of points and objects. 
%------------------------------------------------------------------------
%------------------------------------------------------------------------
\begin{figure*}
\begin{center}
% \fbox{\rule{0pt}{2in} \rule{.9\linewidth}{0pt}}
\includegraphics[width=1\linewidth]{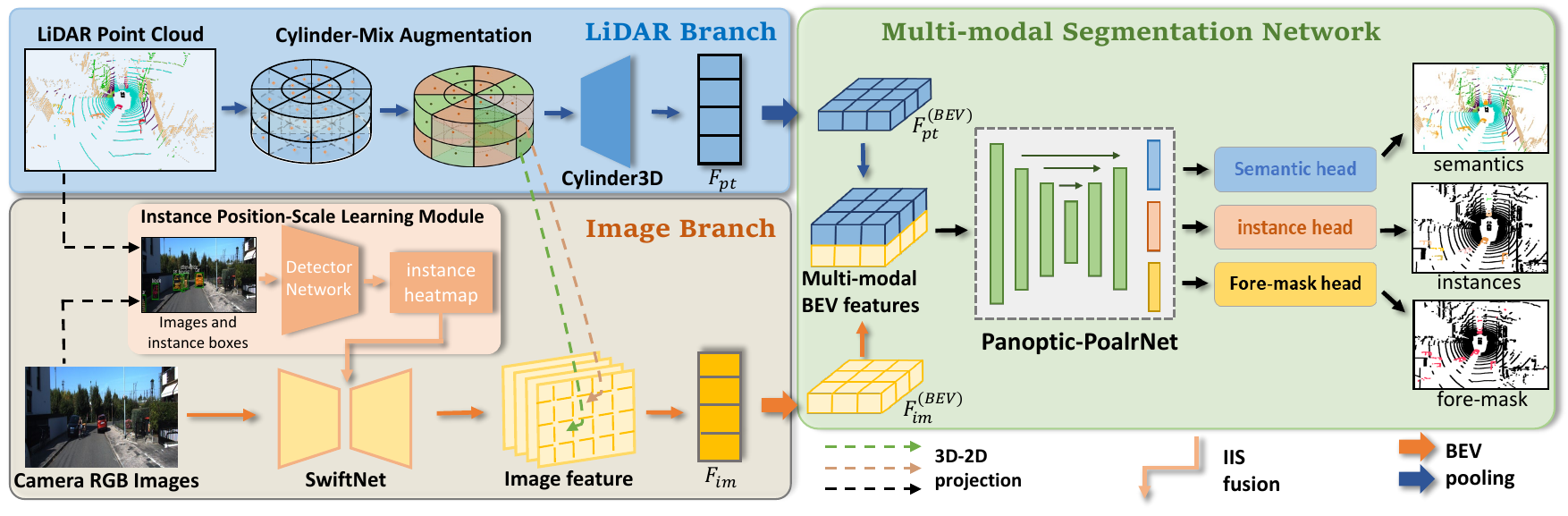}
\caption{The framework of our model. Our model is composed of three parts. LiDAR Branch, on the 3D point cloud branch, gets better 3D features through self-supervised augmentation, called Cylinder-Mix, while Image Branch improves the 2D backbone via fusion of instance position and scale information. After that, both cross-modal features will be fused to the BEV feature, following Multi-modal Segmentation Network to extract features and get point-wise labels in the end.}
\label{framwork}
\end{center}
% \vspace{-2mm}
   % \caption{The framework of .}
\end{figure*}
%------------------------------------------------------------------------

\section{Preliminaries}
%method的行文安排，我们将在sec1... sec2...

% To further elucidate, we first introduce our problem setting and unified symbols, then detail the description of Image Branch, LiDAR Branch and segmentation stream, and the overall training processes. To our best knowledge, our model is the first multi-modal model applying the semi-supervised setting to solve point cloud panoptic segmentation.
%------------------------------------------------------------------------

%\subsection{Semi-supervised Setting}

\textbf{Problem Setting}. 
Semi-supervision usually uses the whole points of partial samples or frames. We adopt a sampling method of selecting frames with fixed intervals, which is the same as \citet{kong2023lasermix}. In this paper, we have respectively selected 40\%, 20\%, 10\%, and 1\% point cloud frames with point-wise labels for training.

% which is also a practical approach for real-world applications
% Furthermore, to compare how many labeled annotations are needed, the percentage of point cloud frames used is helpful to distinguish. Given the percentage, we adopt a sampling method of selecting frames with fixed intervals in our setting, which is also a practical approach for real-world applications. In addition, the percentages of setting are 40\%, 20\%, 10\%, 1\%. 
% Generally, there are three types of ways to get the selected frame when given a percentage, that is random frames of the whole, random scenes of total scenes, and frames with fixed intervals.

% To reduce reliance on point cloud manual annotation, weakly supervised and semi-supervised methods are naturally applied. During training, the former tends to use the partial points on the whole point clouds, while the latter usually uses the whole points of partial point clouds. Furthermore, to compare how many labeled annotations are needed, the percentage of point cloud frames used is helpful to distinguish. Generally, there are three types of ways to get the selected frame when given a percentage, that is random frames of the whole, random scenes of total scenes, and frames with fixed intervals. From a practical point of view, our semi-supervised setting only refers to the latter two. In addition, the percentages of setting are 40\%, 20\%, 10\%, 1\%.

%\subsection{unified Symbols}

\textbf{Symbols}. We use $\textbf{P}=\{(x, y, z, b)\}_{n=1}^N$ to denote a 3D LiDAR point cloud, 
with the RGB image(s) $\textbf{I}=\{I \in \mathbb{R}^{H\times W \times 3}\}_{1}^v$, where $\mathit{(x,y,z)}$ indicates the spatial coordinates, $\mathit{b}$ is the brightness of each point, $N$ and $\mathit{v}$ are respectively the number of points and camera views. 
Besides, there are point-wise semantic and panoptic labels $\textbf{L}=\{\mathit{l}_{sem}, \mathit{l}_{inst}\}_{n=1}^N$, $\mathit{l}_{sem} \in [1,2,3,...,C]$, where C is the number of semantic categories. Therefore, the dataset could be represented as $\mathbb{P}=\{\textbf{P}, \textbf{I}, \textbf{L}\}_{i=1}^M$ with $\mathit{M}$ frames. During training, the dataset is divided into labeled and unlabeled parts, denoted by the superscript $l$ and $u$, respectively. For example, $\textbf{P}^u$ means unlabeled points of the dataset.

\section{Method}

%\subsection{Overview}
%稍微细致的流程，以及部分模块的使用原因和优势
%为了在有限的标注数据中使模型获得更好的分割效果，我们提出的模型在2D图像和3D点云上均有数据上的自监督信息的利用。
%模型同时对图像和点云数据进行处理。一方面，2D图像输入至Image Branch来提取点级的图像特征，其中我们使用了self-info fusion module来添加和融合额外的自监督信息，由3D point投影至image形成的box，接着根据3D-2D的投影坐标获得点的每个point的image 特征。另一方面，LiDAR分支输入3D点云，经过我们提出的Cylinder-Mix数据增强式以及MLP-based网络后，输出point-wise特征。接下来，融合图像和点云的BEV池化特征，输入到分割网络中，经过mutli-heads，我们可以得到点击的语义标签和全景分割(通过聚类)
As shown in Figure \ref{framwork}, 
% our model consists of two branches. Given the input of points and images, the two branches are employed to extract features of different modalities. 
%After fusion of the both branches, the multi-modal features are subsequently fed into a segmentation network for the final panoramic results.
% As illustrated in Figure \ref{framwork}, the framework of our model is a double-branch and two-stage structure. With the input of points and images, the two branches are employed to extract features of different modalities. After fusion of the both, the multi-modal features are subsequently fed into a point cloud segmentation network for panoramic results.
LiDAR Branch, given the labeled point clouds, outputs point-wise features through Cylinder-Mix augmentation and Cylinder3D \cite{zhu2021cylinder3d} extractor network.
On Image Branch, the proposed Instance Position-Scale Learning (IPSL) module outputs instance heatmaps. To this end, a detector, supervised by the 3D-2D projected instance boxes, is introduced into this module to predict boxes. Ultimately, the instance heatmaps are fused into the backbone, SwiftNet \cite{orsic2019swiftnet}, so that the pixel features corresponding to the 3D points on the image are generated.
% On Image Branch, the proposed Instance Position-Scale Learning (IPSL) module outputs instance boxes and is supervised by the latent labels, which are the instance boxes obtained from the point labels through 3D-2D projection. The predicted boxes are encoded into instance heatmap to fuse into the backbone, SwiftNet \cite{orsic2019swiftnet}. Finally, pixel features corresponding to the 3D points on the image are generated.
% \td{the latent labels, the projected instance boxes are obtained from the point labels through 3D-2D projection. These boxes are learnt and predicted on Instance Position-Scale Learning module and then encoded into instance heatmap to fuse into the backbone, SwiftNet \cite{orsic2019swiftnet}. Finally, pixel features corresponding to the 3D points on the image are output.}}
Then, the features from two branches are processed by Multi-modal Segmentation Network, which consists of the BEV-pooling \cite{li2022bevdepth} and Panoptic-PolarNet \cite{zhou2021panoptic-polarnet}.
Finally, we get semantic and instance predictions for each point via multi-heads and clustering. Additionally, self-training is utilized for unsupervised data, similar to \citet{xu2021semi-self-training}.
% , the trained model can be used to estimate the unlabeled inputs and the final model is retrained by the labeled and estimated labels.

% In the later experiments, our baseline consists of an image encoder and a LiDAR encoder (Image Branch and LiDAR Branch pipeline), and a decoder (Multi-modal Segmentation Network), excluding the proposed IPSL module and Cylinder-Mix.

% More specifically for the two branches, our proposed model utilizes self-supervised information both on 2D images and 3D point clouds modalities to achieve more effective segmentation. One hand, point-wise pixel features is extracted on the \textbf{Image Branch}, in which we leverage Instance Position-scale Learning Module to add and fuse additional self-supervision, the 2D boxes got from points at zero cost. On the other hand, with input of point cloud, the \textbf{LiDAR Branch} outputs point-wise features through our Cylinder-Mix augmentation data enhancement and Cylinder3D \cite{zhu2021cylinder3d} extractor network. In the following Multi-modal segmentation, the both features are fused after BEV-pooling, and then sent to Panoptic-PolarNet \cite{zhou2021panoptic-polarnet}. 
% When it comes to the two-stage, during the first phase, called \textbf{pretrain}, our model is trained on labeled data. While in the second stage, the model is \textbf{retrain}ed using both labeled data and unlabeled data with pseudo labels \textbf{estimate}d by the pretrained model. 

%------------------------------------------------------------------------
\subsection{Training LiDAR Branch}

\subsubsection{\textbf{Pipeline}} 
\label{section of LiDAR Branch}

%这一模块不仅继承了cylinder3d网络良好的特征提取能力，还利用了cylinder mix进行监督自增强，进一步增强网络的模型表达能力。
%为了能区别多样的点云特征，我们首先对输入点云p进行paste或pups的数据增强，然后进行cylinder体素化。此外，我们提出了一直新的数据增强方式，which契合cylinder体素化。这些点输入到MLP类似的网络中，得到点击特征，进行BEV池化，获得point BEV feature.
The objective of this branch is to learn well-behaved feature comprehension of Cylinder3D. It also employs Cylinder-Mix for achieving more latent samples and self-supervised enhancement.
%The LiDAR Branch not only inherits well-behaved feature comprehension of Cylinder3D \cite{zhu2021cylinder3d}, but also employs Cylinder-Mix augmentation for self-supervised enhancement and better feature representation.

As shown in the blue part of Figure \ref{framwork}, LiDAR Branch learns point-wise features with the input of a point cloud $\textbf{P}$. Firstly, it voxelized the points into cylindrical voxels with the grid size of $[G_x, G_y, G_z]$ where $G_x$ is the number of blocks of voxels to be split along the X-axis, so as to $G_y, G_z$. This way we get the grid index $(v_x,v_y,v_z)$ that denotes which cylinder-voxel the point $(x,y,z)$ belongs to. Secondly, it applies our proposed Cylinder-Mix to increase the diversity of the point clouds within the limited data. 
%The details of this data augmentation is present in the subsequent section.
%Section \ref{subsection of Cylinder-Mix}. 

% As shown in Figure \ref{framwork}, the backbone of the LiDAR Branch is Cylinder3D \cite{zhu2021cylinder3d}. With the input of a point cloud $\textbf{P}$, LiDAR Branch is excepted to extract points bird's eye view (BEV) feature. Firstly, We voxelize the points to cylinder with the grid size of $[G_x, G_y, G_z]$ where $G_x$ is the number of blocks of voxels to be split along the X-axis, so as to $G_y, G_z$. And thus we get the grid index $(v_x,v_y,v_z)$ that denotes which cylinder-voxel the point $(x,y,z)$ belongs to. Secondly, we apply the Cylinder-Mix data augmentation which could improve the network performance, since it increases the diversity of the point clouds within limited data. The details of this data augmentation we will present in Section \ref{subsection of Cylinder-Mix} 

The augmented points $\textbf{P}_{mix}$ are fed into the MLP-based feature extractor of Cylinder3D, thus getting point-wise features $F_{pt} \in \mathbb{R}^{N \times C_b}$. Finally, through the grid index of points and BEV pooling, we aggregate $F_{pt}$ into LiDAR BEV features $F^{(BEV)} _{pt}$ with the size of ${G_x \times G_y \times G_z \times C_b}$, where $C_b$ is the number of channels of the BEV feature. When pooling the voxel at position $(v_x,v_y,v_z)$, its BEV feature is the maximum of point features $F_{pt}(p)$ within the point set $\textbf{S}_{(x,y,z)}$ in which all points belong to this voxel.
\begin{equation}
    F^{(BEV)} _{pt}(v_x,v_y,v_z) = \mathop{\max}_{p \in \textbf{S}_{(x,y,z)}} {F_{pt}}(p).
\end{equation}
%------------------------------------------------------------------------
\subsubsection{\textbf{Cylinder-Mix Augmentation}}
\label{subsection of Cylinder-Mix}
%In the setting of semi-supervised learning, only partially labeled data is available.
To obtain more data for network training, we propose a point cloud mixture data augmentation, Cylinder-Mix.
%Inspired by LaserMix \cite{kong2023lasermix}, Cylinder-Mix is specifically designed for cylinder-voxel. 
It, without additional annotations, provides more diverse point clouds for network training and adapts well to the feature extraction of Cylinder3D.
% and performs well on real-world datasets.

\begin{figure}[t]
\begin{center}
% \fbox{\rule{0pt}{2in} \rule{0.9\linewidth}{0pt}}
\includegraphics[width=0.65\linewidth]{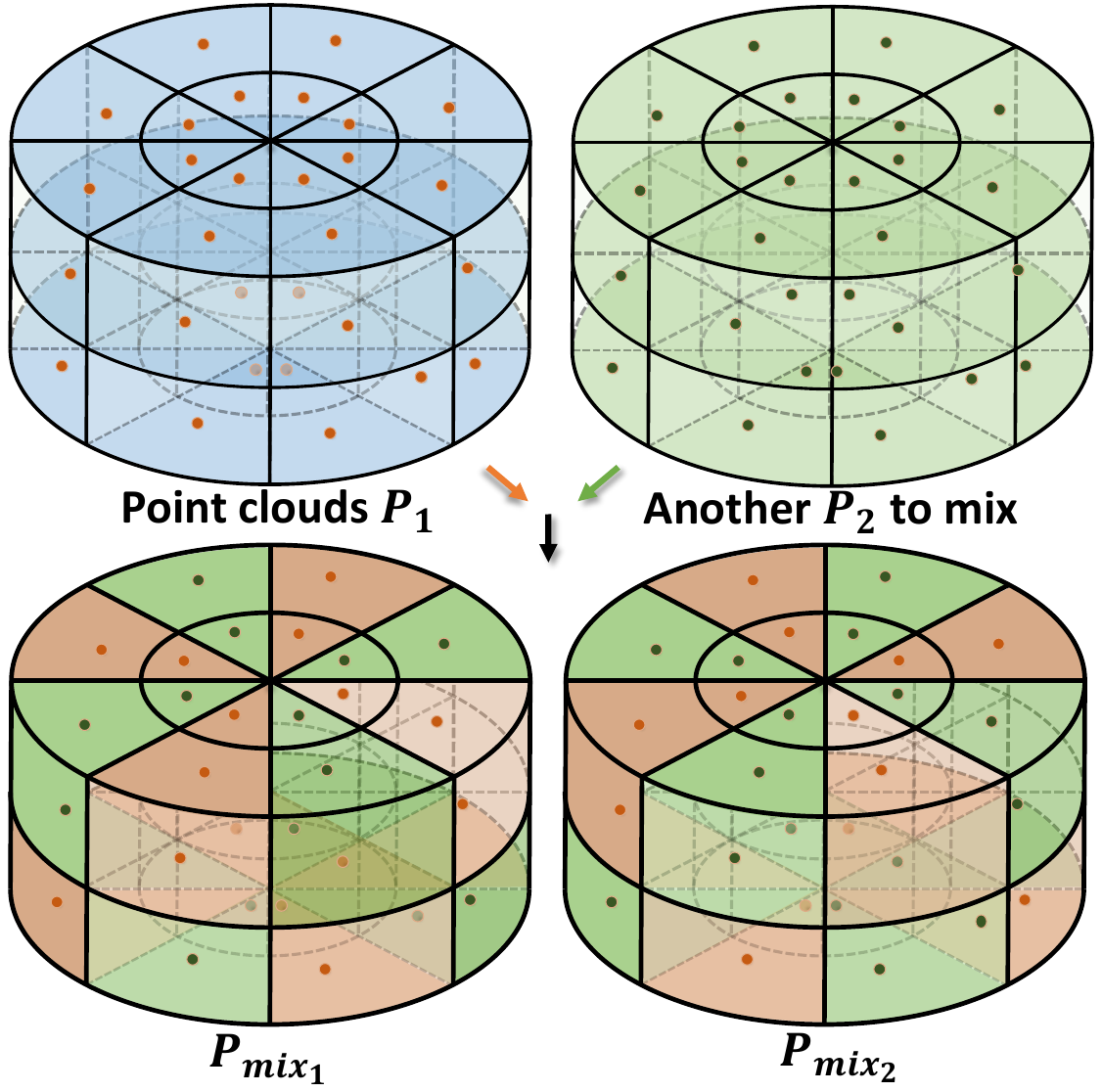}
\end{center}
   \caption{Sketch map of Cylinder-Mix.}
\label{Cylinder-Mix}
% \vspace{-2mm}
\end{figure}

To fully mix the point cloud samples, we perform an interleaved mixing as figure \ref{Cylinder-Mix}.  This means that for a certain block region of the mixed point cloud, its surrounding regions (i.e., above, below, to the left, and to the right) all come from another different point cloud.

Given the point cloud samples $\textbf{P}_1, \textbf{P}_2$ and their cylinder voxel index of each point $V=\{(v_x,v_y,v_z)\}_{n=1}^N$ with the grid size of $[G_x,G_y,G_z]$, we first divide the two point clouds into mixture regions with the size $[R_x,R_y,R_z]$ along each $x,y$ and $z$ axis. 
For example, the region size of the mixture shown in Figure \ref{Cylinder-Mix} is $[8,1,2]$. To mix them, we calculate the $(r_x,r_y,r_z)$ of each point, where $r_x \in \{0,1,2,...,R_x - 1\}$ represents the index number of the mixture region, which point belongs to the $x$ axis, similarly for $r_y$ and $r_z$.
% we firstly voxelize with cylinder grid size  and calculate the voxel index of each points .
% With the input of point $P$ and the cylinder grid size of $[S_x, S_y, S_z]$, , $v_k \in\{0,1,2,...,S_k\}  (k=x,y,z)$. 
\begin{equation}
\label{range block}
    r_k = \lfloor v_k / G_k \times R_k \rfloor, \ \quad (k=x,y,z).
\end{equation}

For interlaced mixing, the two point clouds $\textbf{P}_1, \textbf{P}_2$ are mixed according to index-judgment as the following:
% \begin{equation} 
% \label{mixed index}
% \begin{split}
% \delta(r_k)=\left\{
% \begin{aligned}
% \text{True} & , &{\rm if} \;\ mod(r_k,2)=0 \\
% \text{False} & , &{\rm if} \;\ mod(r_k,2)=1
% \end{aligned}
% \right., (k=x,y,z). \\
% \mathcal{J}(x,y,z) = \neg \left(\delta(r_x) \oplus \delta(r_y)\right) \oplus \delta(r_z).
% \end{split}
% \end{equation}
\begin{equation} 
\delta(r_k)=\left\{
\begin{aligned}
\text{True} & , &{\rm if} \;\ mod(r_k,2)=0 \\
\text{False} & , &{\rm if} \;\ mod(r_k,2)=1
\end{aligned}
\right., (k=x,y,z). \\
\end{equation}
\begin{equation}
\mathcal{J}(x,y,z) = \neg \left(\delta(r_x) \oplus \delta(r_y)\right) \oplus \delta(r_z).
    \label{mixed index}
\end{equation}
In equation \ref{mixed index}, $\neg$ and $\oplus$ indicate logical negation and xor, respectively. $\mathcal{J}_1 (x,y,z)$ is a judgment function for judging whether the point at $(x,y,z)$ will be mixed into the point clouds $\textbf{P}_{mix_1}$. To get the another mixed point clouds $\textbf{P}_{mix_2}$, the index-judgment function $\mathcal{J}_2 = \neg \mathcal{J}_1$.
\begin{equation}
  \textbf{P}_{{mix}_1} = \textbf{P}_1[\mathcal{J}_1 (\textbf{P}_1)]  \cup  \textbf{P}_2[\mathcal{J}_1(\textbf{P}_2)].
    \label{mixed point cloud}
\end{equation}

On this basis, we get mixed points by equation \ref{mixed point cloud}, as shown in Figure \ref{Cylinder-Mix}. Attention that for the mixed point cloud from $\textbf{P}_1$ and $\textbf{P}_2$, its Image Branch should perform feature extraction and projection for camera images $\textbf{I}_1$ and $\textbf{I}_2$, respectively.
In addition, consistently training with mixed data can potentially cause the model distribution to deviate from that of training with original data. Hence, we set $p_{clymix}$, the probability of using Cylinder-Mix, to control this augmentation.

%------------------------------------------------------------------------

%------------------------------------------------------------------------
\subsection{Training Image Branch}
% qianzaid information 多种format
\subsubsection{\textbf{Pipeline}}
\label{section of Image Branch}
Image Branch is aimed at extracting image features and learning instance information from some weak labels. As shown in the orange part of Figure \ref{framwork}, Image Branch consists mainly of two parts: the backbone network (SwiftNet \cite{orsic2019swiftnet} in our experiment) and the Instance Position-scale Learning (IPSL) Module. Further details on this module will be presented in the next section. Given the corresponding images of the point cloud, this branch is expected to output camera BEV features.

% The IPSL Module is aimed to produce an instance heatmap to assist the backbone network. Further details on this module will be presented in the next section.
%The instance information is subsequently fused into the backbone. 

%\ref{section of Instance Position-scale Learning Module}.

%The backbone of Image Branch is SwiftNet \cite{orsic2019swiftnet}. 
With the input images $\textbf{I}$ of point clouds and the fusion of the instance information from the IPSL Module, SwiftNet encodes initial pixel-wise features $F_{im}$.
% However, as these features are still in 2D image coordinates, a correspondence between 3D points and 2D pixels is required
To obtain camera BEV features, we employ the LiDAR-to-camera projection, through camera intrinsic parameters and vehicle parameters, to project the point $p=(x,y,z)$ onto the camera coordinate system $q=(h,w)$. We obtain a set of matched pairs mapping $\textbf{M}=\{p,q\}$ between 3D points and 2D pixels. Subsequently, we perform BEV pooling on the matched pixel features to get the camera BEV feature. Its value on $(v_x,v_y,v_z)$ is the maximum of pixel features $F_{im}(q)$, where $q$ is in set $\textbf{M}_{(v_x,v_y,v_z)}$ and the matched point belongs to its voxel.
\begin{equation}
    F^{(BEV)} _{im}(v_x,v_y,v_z) = \mathop{\max}_{(p,q) \in \textbf{M}_{(v_x,v_y,v_z)}} {F_{im}}(q).
\end{equation}

%------------------------------------------------------------------------

\subsubsection{\textbf{Instance Position-scale Learning Module}}
\label{section of Instance Position-scale Learning Module}
The Instance Position-scale Learning Module is aimed to utilize weak labels on images to provide instance information for improving point cloud segmentation.
Inspired by interactive image segmentation (IIS) \cite{sofiiuk2022reviving_IIS}, images with weak annotations, such as a point indicating the center of an object, make the model pay more attention to the annotated objects. Next, we will introduce how to generate and utilize weak annotations.
% , without additional annotation costs, 

Initially, to train with weak labels, it is necessary to get the latent ground-truth through the instance labels of points and LiDAR-to-camera projection. This involves matching and projecting the 3D points onto 2D pixels. For a specific object with the instance label $y_{inst}$, its box can be enclosed by the maximum and minimum values of pixel coordinates $(h, w)$ from projected points of this instance. Subsequently, we preserve these box labels and use them as targets to train the Detection Network. The Detection Network is an offline network, and it is convenient to adopt any detection backbone. In our experiment, we directly select the detection model from MMdetection \cite{chen2019mmdetection}.

In this way, the module predicts the instance bounding boxes of cameras through the trained Detection Network. Since these boxes are in 2D coordinates, we convert them into instance heatmaps using the following equation:
\begin{equation}
\label{d_mn}
    d_q (m,n) = \sqrt{(h-m)^2 + (w-n)^2},
\end{equation}
% \mathcal{G}
\begin{equation}
H_{q}\left(m,n\right)=\left\{\begin{array}{cl} 
\exp \left( -{{2d_{mn}}^2} / {R^2} \right) & d_{m n} \leq R \\
0 & d_{m n}>R 
\end{array}\right. ,
    \label{heatmap of a point}
\end{equation}

\noindent where $d_q (m,n)$ is the distance on the image coordinate system between pixel $(m,n)$ and $q=(h,w)$. The $H_{q}$ is a Gaussian-based heatmap of a box angular or box center point $q$, with the radius of $R$.

In implementation, the heatmap of an instance is gathered by its 4 box angular points $q_1,q_2,q_3,q_4$ and its center point $q_{cen}$ of the box. $R=5$ for each of the 4 corners, while for its center, $R$ is determined by the product of the instance width and the percentage $P_{center}$. Here, instance width is the minimum value between the box length and width. We set $P_{center}=1/4$ as the default. To ascertain the optimal configuration of these parameters, we conduct comprehensive ablation experiments, as detailed in the Appendix.

With stacking the heatmaps $[H_{q_1},H_{q_2},H_{q_3},H_{q_4},H_{q_{cen}}]$ and maximizing at the stacked dimension, we get the instance heatmap of the box. Furthermore, for the camera image from view $v$, its instances heatmap is the maximum of Gaussian heatmap of each instance.

Finally, referring to the fusion block of IIS \cite{sofiiuk2022reviving_IIS}, these heatmaps and images along with each camera view, are fused together on intermediate features. with camera images $I$. As shown in equation \ref{fusion of image and heatmap}, images $I$ and heatmaps $H$ are aligned through their respective mapping heads $\psi_H$ and $\psi_I$, 2D convolution with the number of output channels 64. And then they are summed to obtain fused intermediate features $F_{IH}$ and fed into the backbone of Image Branch for further extraction.
\begin{equation}
    F_{IH} = \psi_H(\mathcal{H}) + \psi_I(I).
    \label{fusion of image and heatmap}
\end{equation}

% \vspace{+1mm}
\textbf{Instance Heatmap from SAM Masks.} There is another method to obtain instance heatmaps. Due to the development of large models such as CLIP \cite{radford2021CLIP}, SAM \cite{kirillov2023segment-anything}, and Grounded-Segment-Anything \cite{GroundedSAM}, instance masks are still available even without fine-tuning autopilot camera images. Therefore, through zero-shot segmentation by Grounded-Segment-Anything, we get $n_i$ instance masks $Mask$ and their scores $Sco$ on an image. The instance heatmap could be the weighted sum of all those masks.
\begin{equation}
    \mathcal{H} = \sum_{i=1} ^{n_i} {Mask_i * Sco_i}.
\end{equation}

%------------------------------------------------------------------------
\begin{figure}[t]
\begin{center}
\includegraphics[width=1\linewidth]{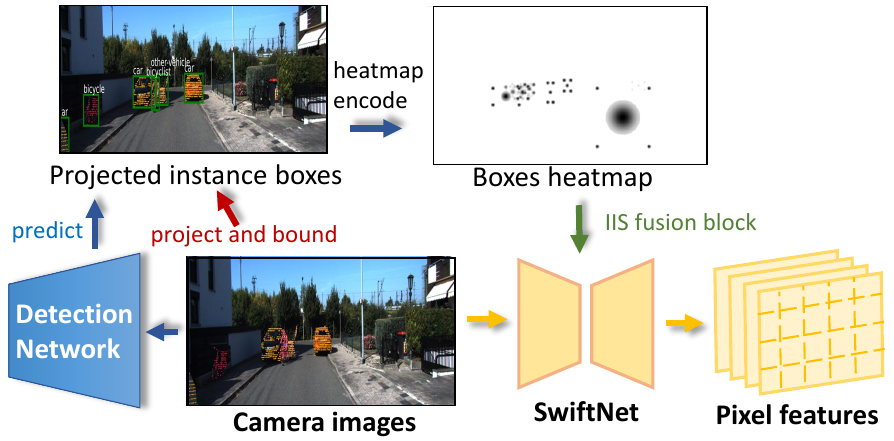}
\end{center}
   \caption{sketch map of IPSL Module on Image Branch.}
\label{image_branch}
% \vspace{-2mm}
\end{figure}
%------------------------------------------------------------------------

\subsection{Multi-modal Segmentation Network}
\label{section of Multi-modal Segmentation Network}
%再获得了多模态的BEV特征后，我们将点云输入到Mult-modal Segmentation Network to 获得语义和全景标签。具体来说，我们通过三个head分别输出语义标签、offset、center heatmap和fore-mask来生成全景标签。
It is expected to get panoptic results with the backbone of Panoptic-Polarnet \cite{zhou2021panoptic-polarnet} and multi-heads.
\begin{equation}
    \label{BEV feature fusion}
    F^{(BEV)} = \mathcal{L}(F^{(BEV)}_{pt} \otimes F^{(BEV)}_{im}).
\end{equation}
The multi-modal BEV features are obtained with LiDAR and Camera ones through equation \ref{BEV feature fusion}, where $\otimes$ means concatenation along the channel dimension and $\mathcal{L}$ represents the linear fully connected layer to compress the feature dimension. $F^{(BEV)}$ is input into the Multi-modal Segmentation Network.
% The multi-modal BEV features are obtained with LiDAR and Camera ones and then input into the Multi-modal Segmentation Network.
To achieve panoramic segmentation, multi-heads include semantic and instance heads, and the latter is responsible for predicting heatmaps of offset and center. Moreover, a fore-mask head is added for \textit{thing} classes perception. We utilize the above mask and heatmaps, based on predicted semantic labels, to cluster and ultimately identify instances.

\textbf{Segmentation Loss.} The loss of segmentation$\mathcal{L}_{seg}$, besides semantic loss and instance losses of the panoptic network, we also added foreground mask loss. With predicted logits of each point on semantic head and ground-truth, we could get semantic loss $\mathcal{L}_{sem}$ by \textbf{cross-entropy} loss. For instance head, we use \textbf{MSE} loss($\mathcal{L}_{hm}$) and L1 Loss($\mathcal{L}_{os}$) to fit the center heatmap of instances $\textbf{hm} \in \mathbb{R}^{H\times W \times 1}$ and offset of each point $\textbf{os} \in \mathbb{R}^{H\times W \times 2}$. For fore-mask head, we use \textbf{MSE} loss($\mathcal{L}_{fm}$)to fit foreground mask $\textbf{fm} \in \mathbb{R}^{H\times W \times 1}$, in which 1 and 0 denote \textit{thing} and background classes. The final loss with coefficients is 
\begin{equation}
    \mathcal{L}_{seg} = \mathcal{L}_{sem} + \mu_{hm} \mathcal{L}_{hm} + \mu_{os}\mathcal{L}_{os} + \mu_{fm}\mathcal{L}_{fm}.
    \label{loss_segmentation}
\end{equation}

%(参考文献，聚类)
%(对于fore-mask的利用需要用公式符号表示)
% ----------------------------------------------------------------
\subsection{Self-Training}
\label{section of self-training}
First, we train our model with labeled data and loss as equation \ref{loss_segmentation}, which is called \textbf{pretraining}. Second, pseudo labels of unlabeled data are \textbf{estimated} with the pretrained model. Third, after loading with the parameter of pretraining, we will \textbf{retrain} model with labeled and estimated data.
Noted that with the decrease in the amount of labeled data during pretraining, the lower semi-supervised percentage, the more epochs are required. And the specific epoch of each setting is given in the section on Experiments. 

\begin{table*}[]
%\vspace{-5mm}
\begin{center}
% \footnotesize
\begin{tabular}{c|l|ccccc|ccccc}
\toprule 
\multirow{2}{*}{}     & \multicolumn{1}{c|}{\multirow{2}{*}{Method}} & \multicolumn{5}{c|}{nuScenes}             & \multicolumn{5}{c}{SemanticKITTI}         \\ 
                      & \multicolumn{1}{c|}{}                        & 1\%   & 10\%  & 20\%  & 50\%*     & 100\% & 1\%   & 10\%  & 20\%  & 50\%*    & 100\% \\ 
\midrule
\multirow{8}{*}{\rotatebox{90}{mIoU}} & CutMix-Seg (BMVC2020)       & 43.8  & 63.9  & 64.8  & 69.8     & /     & 37.4  & 54.3  & 56.6  & 57.6      & /     \\
                      & MeanTeacher (NIPS2019)                      & 51.6  & 66.0  & 67.1  & 71.7      & /     & 45.5  & 57.1  & 59.2  & 60.0       & /     \\
                      & CBST (ECCV2018)                             & 53.0  & 66.5  & 69.6  & 71.6      & /     & 48.8  & 58.3  & 59.4  & 59.7      & /     \\
                      & CPS (CVPR2021)                              & 52.9  & 66.3  & 70.0  & 72.5      & /     & 46.7  & 58.7  & 59.6  & 60.5      & /     \\
                      & LaserMix (CVPR2023)                         & 55.3  & 69.9  & 71.8  & 73.2      & /     & 50.6  & 60.0    & 61.9  & 62.3      & /     \\
                      & baseline                                    & 56.2 & 72.7 & 77.0 & 79.8*    & 80.7 & 51.2 & 61.8 & 62.6 & 64.7*   & 64.2 \\
                      & our model                                   & \textbf{59.1} & \textbf{76.0} & \textbf{78.7} & \textbf{80.5}*  & 82.1 & \textbf{52.8} & \textbf{64.8} & \textbf{64.9} & \textbf{65.9}*    & 65.3 \\
                      & $\Delta \uparrow$                           & +2.9 & +3.3 & +1.7 & +0.7     & +1.4 & +1.6 & +3.0 & +2.3 & +1.2     & +1.1  \\ \hline
\multirow{3}{*}{\rotatebox{90}{PQ}}   & baseline                    & 47.0 & 67.8 & 71.9 & 75.5*    & 77.6 & 46.5 & 57.0 & 59.3 & 60.4*     & 59.8 \\
                      & our model                                   & \textbf{49.8} & \textbf{70.9} & \textbf{75.6} & \textbf{77.2}*     & 78.6 & \textbf{49.0} & \textbf{59.9} & \textbf{60.8}  & \textbf{61.2}*  & 60.8 \\
                      & $\Delta \uparrow$                            & +2.8 & +3.1 & +3.7 & +1.7     & +1.0 & +2.5 & +2.9  & +1.5 & +0.8      & +1.0  \\ 
\bottomrule
\end{tabular}
\end{center}
% \vspace{-3mm}
\caption{Segmentation results compared with other semi-supervised methods on validation set of nuScenes and SemanticKITTI. The symbol * in the table denotes semi-supervised results at a ratio of 40\%, distinguishing them from other methods at 50\%.}
\label{table: segmentation results}
\end{table*}
%------------------------------------------------------------------------

\section{Experiments}
% This section, we introduce the datasets and training details, as well as the main results of performance and visualization. Finally, we conduct thorough ablation study on our model.
%------------------------------------------------------------------------

\subsection{Dataset and Metrics}
% , providing semantic and instance annotations for KITTI \cite{geiger2012kitti} odometry dataset, is the first autopilot dataset for LiDAR panoptic segmentation. SemanticKITTI 
\textbf{SemanticKITTI} \cite{behley2021benchmark} contains 10 (1/11) training (validation/testing) sequences and totally 43551 LiDAR scans with a 64-beam LiDAR sensor, as well as binocular camera images of each scan additionally. There are point-wise panoptic annotations with 8 \textit{thing} class and 12 \textit{stuff} class labels with instance labels. \\
\textbf{nuScenes} \cite{caesar2020nuscenes} is a large-scale autopilot dataset with various urban scenes and a 32-beam LiDAR sensor. It totally contains 1000 driving scenes of 20s duration and point-wise panoptic annotations, with 16 semantic classes, 10 of which are \textit{thing} classes. For the camera, there are 6 views of images per scan.\\
\textbf{Metrics}. Typically, panoptic quality (PQ) is used to evaluate panoptic segmentation as \citet{kirillov2019panoptic}. And mean IoU (mIoU) reflects the quality of semantic segmentation. For the detector on IPSL Module, average precision (AP) is used to measure the quality of predicted boxes.
% SQ and RQ respectively mean the average IoU and the quality of the true positive rate of each predicted and ground-truth instance.

\subsection{Implementation Details} 
\textbf{Our baseline} comprises an image encoder (SwiftNet), a LiDAR encoder (Cylinder3D), and a decoder (Panoptic-PolarNet). It can be comprehended as the combination of section of \textit{Image Branch and LiDAR Branch pipeline}, as well as \textit{Multi-modal Segmentation Network}, with the exclusion of the description of the proposed IPSL module and Cylinder-Mix. Additionally, the instance augmentation of PUPS \cite{su2023pups} is applied.\\
% Additionally, the instance augmentation of PUPS \cite{su2023pups} is applied. More \textit{Implementation Details} of Detection Network and point cloud segmentation network are in the Appendix \cite{chen2023label}.
% \textbf{Cost} On nuScenes, the number of baseline parameters is 296.2MB, and the memory cost is 26.2GB (batch size=2). With Cylinder-Mix (CM), the memory and time cost (1.2 seconds per frame) remain nearly unchanged. When the IPSL module is added, memory and time costs acceptably increase by 3.5GB and 0.13 seconds per frame, respectively.
\textbf{Implementation} Default settings unless ablation. In IPSL Module, the radius of Gaussian heatmaps $R=5$, percentage of scale to the center $P_{center}=1/4$. For Cylinder-Mix, the region size $[R_x, R_y, R_z]=[4,4,2]$, the probability $p_{clymix}=25\%$. For the weight of each term in loss in $\mathcal{L}_{seg}$, $\mu_{hm}=100, \mu_{os}=10, \mu_{fm}=1$. More \textit{Implementation Details} of Detection Network and point cloud segmentation network are in the Appendix.

\begin{figure}
\begin{center}
\includegraphics[width=1\linewidth,height=0.58\linewidth]{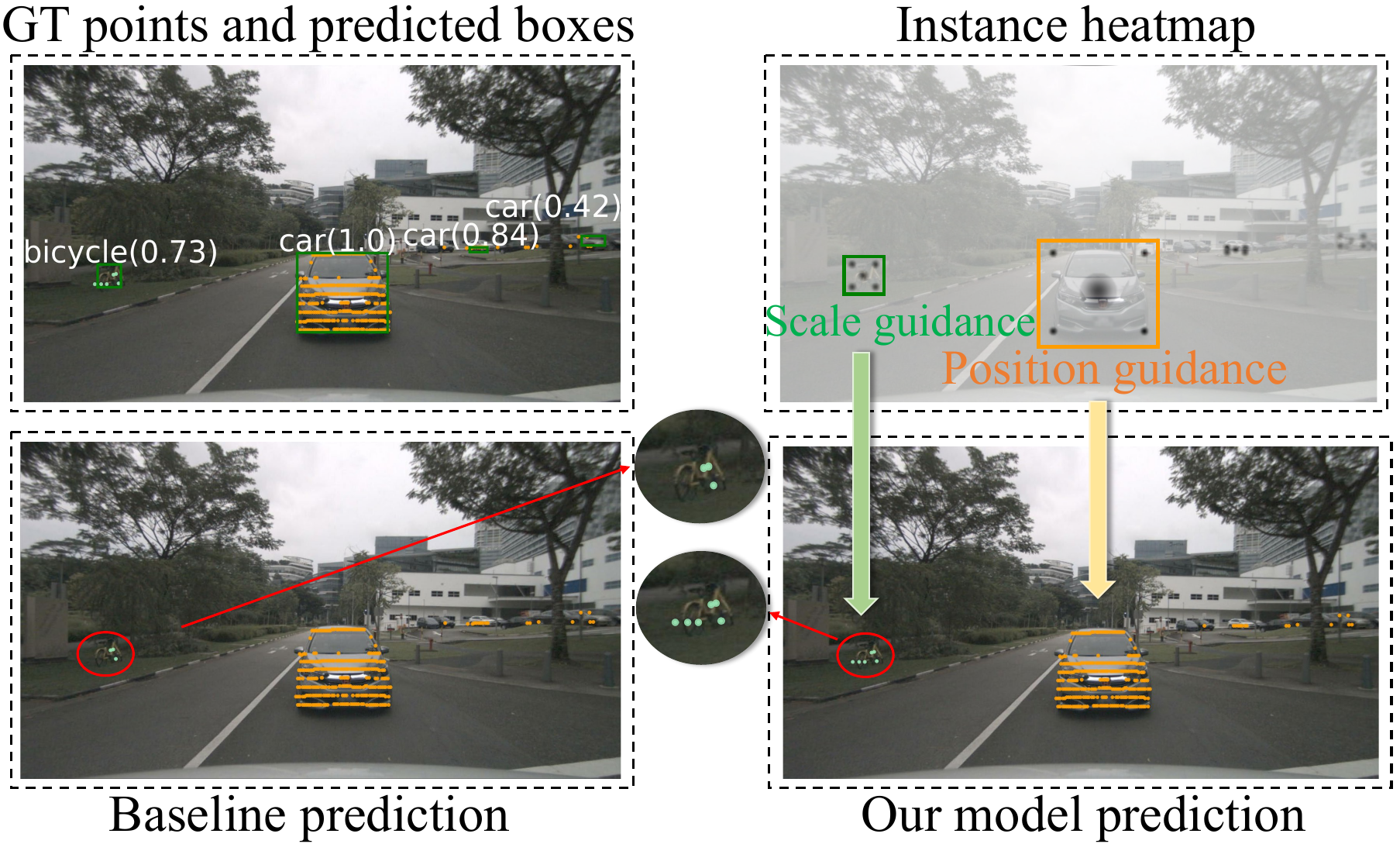}
\end{center}
\caption{Guidance on instance position and scale. After the detector predicts instance boxes, our model generates instance heatmaps (top-right image) to provide position and scale information. See the bicycle (better zoom in), it achieves better segmentation compared to the baseline.}
\label{visualization of position-scale}
% \vspace{-5mm}
\end{figure}

%------------------------------------------------------------------------
\subsection{Experimental Results}
\subsubsection{Results of Detection Network}
We utilize Faster R-CNN with ResNet-101 \cite{he2016resnet} to train the detector. The result mAP on the fully nuScenes dataset is only 31.2\%. However, these predicted instance boxes, acting as weak labels, can enhance point segmentation even without high precision. See the detailed class-wise mIoU in the Appendix.
% There are the average precision(AP) of each \textit{thing} class and mAP (\%) on the kittiImages and nuImages, which refer to the camera images of SemanticKITTI and nuScenes, respectively. The mAP of the Detection Network on both datasets appears to be low, this is due to the imbalance in the number of instances and the confusion of semantics. Nevertheless, its AP could reach 65.9\% if using weighted sum by the number of instances. However, this also indicates the effectiveness of IPSL from another perspective. As these predicted instance boxes, as a form of weak labels, can still improve point cloud segmentation even without high precision.
% We now present the detection network metrics between the ground-truth bounding boxes projected from 3D points and the predicted ones from Detection Network. As shown in \ref{table: results on detection network}, 
% For example, motorcycle and motorcyclist similar in semantic classes, and their positions of ground-truth bounding boxes are also very close. On SemanticKITTI, what’s more there are only 130 instances of motorcyclist, and its AP close to 0, which significantly lower the class average AP. Besides, despite of failing to distinguish semantic classes, it does not affect Instance information of our IPSL module, as the instance heatmap is only related to the predicted box position but not to semantics.
% trained detector

\subsubsection{Results of Panoptic Segmentation} 
\label{section of Results of Panoptic Segmentation} 

Firstly, we present the \textbf{Quantitative Results} as Table \ref{table: segmentation results}. It shows segmentation results on the validation set. As there are limited panoramic segmentation methods available for SS methods, we compare our model with previous methods \cite{french2019cutmix-seg,tarvainen2017mean-teacher,zou2018CBST,chen2021CPS,kong2023lasermix} with semantic segmentation metrics (mIoU, \%) and attach our panoptic segmentation metrics (PQ, \%). Eventually, our model outperforms the previous top, LaserMix, and attains state-of-the-art semantic results. 

Our model excels in the semi-supervised setting, as it exhibits superior performance when given a smaller amount of labeled data. On nuScenes and SemanticKITTI at 10\%, the model achieves 3.1\% and 2.9\% improvement on PQ and 3.3\% and 3.0\% improvement on mIoU, respectively, compared to the baseline. Moreover, at the setting of 20\% labeled data on SemanticKITTI, our model achieves comparable performance to the fully-supervised baseline, and even surpasses that at the setting of 40\%.
% there is a trend that the lower the proportion of semi-supervision, the greater the improvement.
% , since it is not fully trained and using Cylinder-Mix is equivalent to obtaining various data to train. Furthermore, image features are more easily learned compared to point clouds under the same amount of limited data, thus the Instance Position-scale Learning Module plays a larger role.

% Additionally, it may be unfair to compare the semantic segmentation model with our model which uses instance labels, hence we designed another experiment to estimate this additional benefit from instance information. We remove the instance-related predictions from the multi-Heads and only utilize the semantic head for training. Comparing the mIoU from semantic segmentation networks and panoptic segmentation networks on our baseline, we found that the additional mIoU is usually 1.5\% and 1\% on nuScenes and SemanticKITTI, respectively. In other words, even if the additional improvement from the instance label is ignored, our model still outperforms the LaserMix. 

%------------------------------------------------------------------------

\begin{figure}
% \fbox{\rule{0pt}{2in} \rule{.9\linewidth}{0pt}}
% \includegraphics[width=0.9\linewidth]{samples/pics/boundary_benifit2.pdf}
\centering
\subfloat{\includegraphics[height=0.172\textwidth]{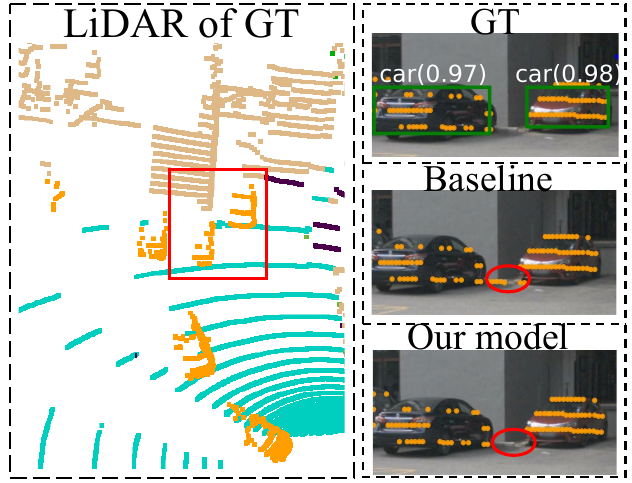}}\hfill
\subfloat{\includegraphics[height=0.172\textwidth]{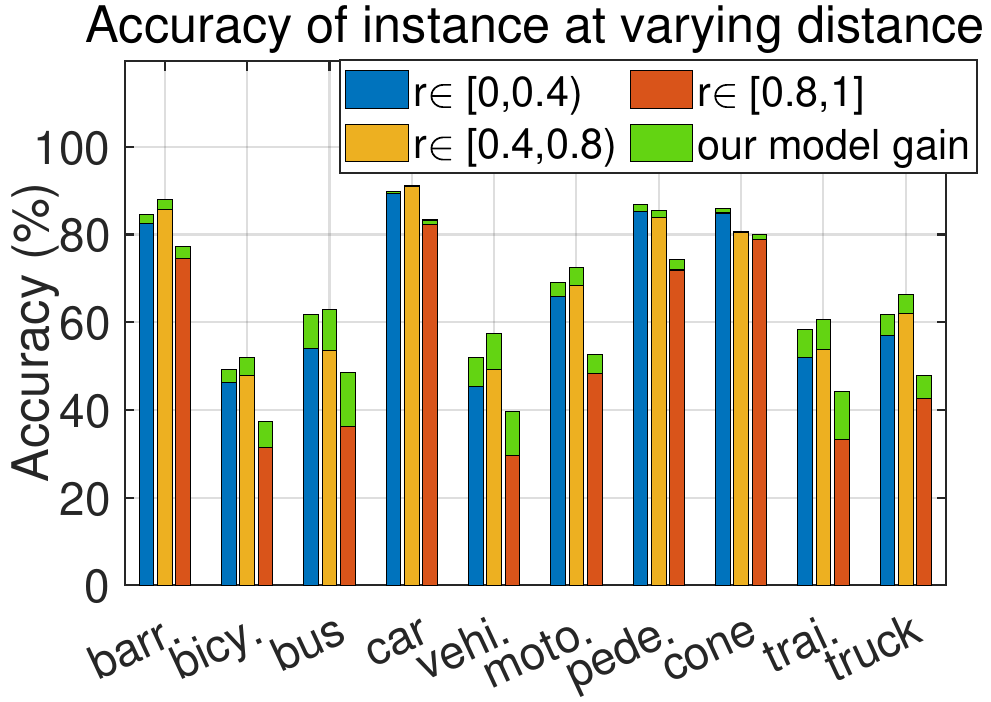}}
\caption{Enhanced performance near the instance boundaries. The left image illustrates the ground-truth and the projected prediction, where our model improves segmentation on the boundaries between instances. The right presents the class-wise classification accuracy. Each set of three bars corresponds to points close, medium, and far from the instance center, based on the distance-to-instance size ratio $r_{ds}$. The upper part represents the enhancement achieved.}
\label{visualization of benefiting boundary}
%\vspace{-5mm}
\end{figure}

%问题：nuscenes数据集中，我们的分割结果对val数据进行了处理，导致pq提高，这个可能需要说明
% , as well as its superior segmentation on instance boundary points.
\textbf{Visualization.} 
Our model excels in excels in \textbf{scale and position guidance}.
As shown in Figure \ref{visualization of position-scale}), with the instance heatmaps, our model achieves superior segmentation in terms of instance scale. Moreover, the heatmap position guidance helps alleviate false positives in some small and distant instances.
Figure \ref{visualization of benefiting boundary} illustrates our model's superior segmentation on \textbf{instance boundary} points. In terms of metrics, the bar graph on the right (representing points distant from the instance center) of each class shows the greatest accuracy increase, suggesting that our model pays more attention to points near the boundaries.

%------------------------------------------------------------------------

%------------------------------------------------------------------------
\subsection{Ablation Studies}
% In this section, we compare our model with LaserMix and then conduct overall analysis of each component to demonstrate their effectiveness. Finally the detailed ablations.

% In this section, we conduct ablation experiments to demonstrate the effectiveness of each component. Firstly, we thoroughly compare our model with LaserMix. Next, we perform an overall analysis of the proposed components, followed by more detailed ablations on each component.

% we first conduct an overall analysis of proposed components and their promotion and then conduct more detailed ablation experiments on each component.

\subsubsection{Compare with LaserMix}
For a fair comparison, we compare our model with LaserMix under the same baseline. As demonstrated in Table \ref{table: LaserMix and Cylinder-Mix}, there are two scenarios: firstly, using LaserMix not only its framework (Mean Teacher) but also its data augmentation simultaneously (baseline + LaserMix), and secondly, employing only the augmentation while keeping other conditions constant (baseline + LaserMix*).

% \textbf{{Outperform LaserMix.}} 
In the third row of Table \ref{table: LaserMix and Cylinder-Mix}, LaserMix exhibits improvements at 10\% and 20\% semi-supervised ratios. Nevertheless, our Cylinder-Mix surpasses it. Second, to further eliminate the framework bias, we exclusively use LaserMix instead of Cylinder-Mix for data augmentation while keeping all other factors unchanged. The third and seventh rows of Table \ref{table: LaserMix and Cylinder-Mix} reveal that Cylinder-Mix consistently outperforms LaserMix.

% Another reason of our model outperforming baseline+LaserMix, is providing more reliable samples than pseudo-labels for training.
\textbf{{Reliable potential samples.}} As shown in the \ref{table: LaserMix and Cylinder-Mix}, the pretrain-retrain approach is employed in LaserMix, where only labeled point clouds are mixed and these samples are reliable. In contrast, the Mean Teacher approach, employed in LaserMix*, mixes labeled and unlabeled points, with pseudos predicted by the Teacher Network. However, these pseudos are unreliable, especially in a low semi-supervised ratio. Both are fundamentally identical, except that the former train has additional and reliable mixed samples derived from labeled point clouds. Experimental results demonstrate that the former outperforms the Mean Teacher approach comprehensively. This indicates that the labels augmented by Cylinder-Mix are reliable and of high quality.

% First, we employed LaserMix not only its framework (Mean Teacher) but also its data augmentation to train our multi-modal panoptic baseline. As the third row of Table \ref{table: LaserMix and Cylinder-Mix}, LaserMix exhibits improvements at the 10\% and 20\% semi-supervised ratios. However, our proposed Cylinder-Mix surpasses the former. Second, to further eliminate the framework bias, we exclusively employed LaserMix as a data augmentation method in place of Cylinder-Mix, while keeping all other factors unchanged. The third and the seventh row of Table \ref{table: LaserMix and Cylinder-Mix} reveal that our model consistently outperforms LaserMix.

%------------------------------------------------------------------------
% baseline+LaserMix  applies LaserMix not only its framework but also its data augmentation to our baseline, while baseline+LaserMix* exclusively substitutes the data augmentation of LaserMix for Cylinder-Mix.
\begin{table}[]
\begin{center}
% \vspace{-5mm}
% \footnotesize
\small
\begin{tabular}{c|l|cccc}
\toprule 
                      & \multicolumn{1}{c|}{}            & 1\%   & 10\%  & 20\%  & 40\%  \\ \midrule
\multirow{4}{*}{\rotatebox{90}{PQ}}   & baseline       & 46.5 & 57.0 & 59.3 & 60.4  \\
                      & baseline+LaserMix              & 46.0    & 56.5 & 59.0 & 59.7 \\
                      & baseline+LaserMix*             & 47.0   & 58.4 & 60.0 & 59.9 \\
                      & baseline+Cylinder-Mix          & 49.0   & 59.9 & 60.8  & 61.2  \\ \hline
\multirow{4}{*}{\rotatebox{90}{mIoU}} & baseline       & 51.2 & 61.8 & 62.6 & 64.7 \\ 
                      & baseline+LaserMix              & 49.7 & 62.3 & 63.9 & 64.8 \\
                      & baseline+LaserMix*             & 50.4 & 64.1 & 64.7 & 65.5 \\
                      & baseline+Cylinder-Mix          & 52.8 & 64.8 & 64.9 & 65.9 \\ 
                      \bottomrule
\end{tabular}
\end{center}
% \vspace{-3mm}
\caption{Comparison of Cylinder-Mix between LaserMix.}
\label{table: LaserMix and Cylinder-Mix}
\end{table}
% ------------------------------------------------------------------------

%------------------------------------------------------------------------
\begin{table}[]
\begin{center}
\begin{tabular}{ccc|ccc}
\toprule
baseline          & IPSL         & CM           & 10\%  & 20\%  & 40\%  \\ \midrule
\checkmark        &              &              & 67.8 & 71.9 & 75.5 \\
\checkmark        & \checkmark   &              & 68.4 & 74.0 & 76.2 \\
\checkmark        &              & \checkmark   & 68.5 & 74.6 & 76.1 \\
\checkmark        & \checkmark   & \checkmark   & 70.9 & 75.6 & 77.2  \\ 
\bottomrule
\end{tabular}
\end{center}
% \vspace{-5mm}
\caption{Ablation of IPSL and Cylinder-Mix on nuScenes}
\label{Table: Ablation experiments on nuScenes}
\end{table}
% %------------------------------------------------------------------------

\subsubsection{Components Analysis}

We conduct ablations on the Instance Position-scale Learning (IPSL) Module and Cylinder-Mix (CM) with their default settings.
% In IPSL Module, the radius of Gaussian heatmaps $R=5$, percentage of scale to the center $P_{center}=1/4$, and the region size and the probability of CM are $[4,4,2]$ and 25\%, respectively.

\textbf{Ablation on nuScenes}. As shown in Table \ref{Table: Ablation experiments on nuScenes}, we presented the PQ values with semi-supervised ratios of 10\%, 20\% and 40\%, where \checkmark indicates whether IPSL and Cylinder-Mix are used. The results indicate that both modules are effective on the nuScenes dataset. IPSL performed best at a 20\% proportion, bringing a 2.1\% increase in PQ, while Cylinder-Mix shows a 2.7\% increase at the 10\% setting. Cylinder-Mix tends to have a greater impact with a smaller amount of data. Moreover, we can draw the conclusion that both the Cylinder-Mix and IPSL module exhibit performance gain on nuScenes dataset.

Ablation on SemanticKITTI is present in the Appendix.

\subsubsection{Instance Position-Scale Learning Module}
% todo 确定2D Detection Network使用半监督还是全监督

We conduct ablation experiments for the IPSL to demonstrate its \textbf{strong generality}, where even when using detectors from different datasets or detectors with slightly lower accuracy on partial datasets, the performance is comparable.   

\textbf{Undemanding Detector}. First, IPSL does not require high accuracy from the detector. As shown in Table \ref{table: universality of IPSL}, comparing IPSL(SS-detector) and IPSL(fully detector), the mAP of the detector trained with a partial dataset is slightly lower than that of the fully trained detector by 31.2\%. However, their PQ results are comparable. 
% indicating that IPSL does not require high precision from the Detector Network. 
% Moreover, even with a semi-supervised detector, our method still outperforms other approaches, as demonstrated in Table \ref{table: segmentation results}.

\textbf{{From Different datasets.}} Second, IPSL still works with detectors from different datasets. Due to the class-agnostic nature of the heatmap in IPSL, the detector trained on SemanticKITTI can be used to predict boxes for nuScenes images. The similarity between IPSL(KITTI-detector) and IPSL(fully detector) in Table \ref{table: universality of IPSL} further confirms the universality of the detector of IPSL.

\textbf{{From LLM masks.}} Third, comparing IPSL (GSA mask) with the baseline and IPSL (fully detector), we observe that utilizing masks predicted by a large model, Ground-Segment-Anything (GSA), remains effective in IPSL. This further validates the generality of the IPSL module and aligns with our initial motivation for designing IPSL.

% the IPSL module is designed to improve semi-supervised segmentation by reinforcing the 2D branch, cause images are cost-effective and easier to learn compared to point clouds. Moreover, with the development in large-scale image models, it is imperative to leverage their advantages in semi-supervised and unsupervised learning to improve the point clouds perception. Therefore,

%------------------------------------------------------------------------  
\begin{table}[]
\begin{center}
\begin{tabular}{l|ccc}
\toprule 
                                     & 10\%       & 20\%    & 40\%  \\ \midrule
baseline                             & 67.8      & 71.9   & 75.5 \\ \hline
\quad mAP of SS-detector             & 23.8       & 27.2    & 28.1  \\
\quad+ IPSL(SS-detector)             & 70.2      & 75.1   & 76.3 \\ \hline
\quad+ IPSL(KITTI-detector)          & 69.9      & 74.8   & 76.0 \\ \hline
\quad+ IPSL(GSA mask)                & 70.7       & 75.3    & 76.3 \\ \hline
\quad+ IPSL(fully detector)        & 70.9      & 75.6   & 77.2  \\ \bottomrule
\end{tabular}
\end{center}
\caption{Ablation study of IPSL on nuScenes. SS-detector and fully detector are trained with semi-supervised and fully nuScenes, while KITTI-detector from SemanticKITTI.} 
% SS-detector: semi-supervised detector on nuScenes. fully detector: trained from fully nuScenes. detector KITTI-detector: detector trained with SemanticKITTI.
% The part in parentheses (*) after IPSL indicates the source of the heatmap: either the predicted boxes from the trained detector or the GSA masks from the larger model GSA, (semi-supervised) SS-detector, fully (nuScenes) detector, and KITTI-detector.}
% refer to the Detector Network trained with semi-supervised and fully supervised data on nuScenes, as well as fully SemanticKITTI, respectively
% "SS-detector" is the Detector Network trained with semi-supervised data, while "fully detector" is trained with fully one. 
\label{table: universality of IPSL}
% \vspace{-5mm}
\end{table}

\subsubsection{Fine-grained Analysis of Module Parameters}
We perform ablations on the parameters of the IPSL module and Cylinder-Mix, including $P_{center}$ for IPSL and $p_{cylmix}$ and region size for Cylinder-Mix, to ensure their rationality. Detailed results and analyses can be found in the Appendix.

\section{Conclusion}
In this paper, we exploit the latent labels from the original labels for semi-supervised multi-modal point cloud panoptic segmentation. For LiDAR data, we introduce a novel data augmentation to generate more and reliable point clouds from limited labeled data. For images, we propose the IPSL module to learn and fuse the information of instance position and scale. We compare our approach with other methods and demonstrate the effectiveness and adaptability of our model by applying a large-scale architecture to our proposed module. Our approach provides inspiration for understanding and mining deeper information from LiDAR-Camera data.

\section{Acknowledgments}
This work is supported by the National Natural Science Foundation of China (No.62222602, 62302167, U23A20343, 62106075, 62176224,62176092),  Development Project of Ministry of Industry and Information Technology (ZTZB-23-990-016),  Shanghai Sailing Program under Grant (23YF1410500), Natural Science Foundation of Shanghai (23ZR1420400), Natural Science Foundation of Chongqing (CSTB2023NSCQ-JQX0007, CSTB2023NSCQ-MSX0137), CCF-Tencent Rhino-Bird Young Faculty Open Research Fund (RAGR20230121), CAAI-Huawei MindSpore Open Fund.

% \begin{itemize}
%     \item Anonymous submissions must not include the author names and affiliations. Write ``Anonymous Submission'' as the ``sole author'' and leave the affiliations empty.
%     \item The PDF document's metadata should be cleared with a metadata-cleaning tool before submitting it. This is to prevent leaked information from revealing your identity.
%     \item References must be anonymized whenever the reader can infer that they are to the authors' previous work.
%     \item AAAI's copyright notice should not be included as a footer in the first page.
%     \item Only the PDF version is required at this stage. No source versions will be requested, nor any copyright transfer form.
% \end{itemize}

\bibliography{references}

% -----------------------------------------------------------------------------------------

\clearpage
% \title{Appendix}
% \maketitle
% \appendix
\twocolumn[
\begin{center}
    {\LARGE \textbf{Appendix}}
    \vspace{5mm}
\end{center}
]

\appendix
\setcounter{figure}{0}
\setcounter{table}{0}

\section{Detailed Experiments}

% \subsection{Dataset and Metrics}
% \textbf{SemanticKITTI} \citet{behley2021benchmark}, providing semantic and instance annotations for KITTI \cite{geiger2012kitti} odometry dataset, is the first autopilot dataset for LiDAR panoptic segmentation. SemanticKITTI contains 10 (1/11) training (validation/testing) sequences and total 43551 LiDAR scans with a 64-beams LiDAR sensor, and binocular camera images of each scan additionally. There are point-wise panoptic annotations with 8 \textit{thing} class and 12 \textit{stuff} class labels with instance labels. 

% \textbf{nuScenes} \cite{caesar2020nuscenes} is a large-scale autopilot dataset with various urban scenes and a 32-beams LiDAR sensor. It totally contains camera 1000 driving scenes of 20s duration and point-wise panoptic annotations, with 16 semantic classes, 10 of which are \textit{thing} classes. For the camera, there 6 views of images per scan.

% \textbf{Metrics}. Usually, panoptic quality (PQ) is used to evaluate panoptic segmentation as \citet{kirillov2019panoptic}. SQ and RQ respectively mean the average IoU and the quality of the true positive rate of each predicted and ground-truth instance. Additionally, mean IoU (mIoU) is used to reflect the quality of semantic segmentation. For the detector on Instance Position-scale Learning Module, average precision (AP) to measure the quality of predicted boxes.

\subsection{Implementation Details}

\textbf{Point Cloud Panoptic Segmentation}. Following the same configuration of Panoptic-PolarNet, the points among the area of $\{(x,y,z)\mid (3\leq \sqrt{x^2+y^2} \leq 50) \bigcap (-3\leq z\leq 1.5) \}$ are voxelized with the grid size [480, 360, 32]. For Cylinder-Mix, the region size $[R_x, R_y, R_z]=[4, 4, 2]$. the weights of each term in loss $\mathcal{L}_{seg}$, $\mu_{hm}=100, \mu_{os}=10, \mu_{fm}=1$. Besides, we set the learning rate at 0.002 and 0.004 for the datasets of SemanticKITTI and nuScenes, respectively. During the pretraining phase, as discussed in the \textit{Self-Training} section of the paper, we configure 240, 300, 400, and 800 epochs for the experiments of semi-supervised ratios of 40\%, 20\%, 10\%, and 1\% on nuScenes. Similarly, for SemanticKITTI, we set 100, 180, 260, and 680 epochs for the corresponding SS ratios. While the epochs of the retraining phase are always 25.
% we set 240, 300, 400, and 800 epochs for 40\%, 20\%, 10\%, and 1\% semi-supervised ratios on nuScenes and 100, 180, 260, and 680 epochs for the same semi-supervised ratios on SemanticKITTI. 

\noindent \textbf{Detection Network}. The detector is trained on MMdetection \cite{chen2019mmdetection} offline after generating and saving instance boxes in images by 3D-2D projection. We choose Faster R-CNN \cite{ren2015faster} as the backbone and train with a learning rate of 0.002 for SmanticKITTI and 0.005 for nuScenes images. Besides, we set 0.65 as the RPN threshold and 0.45 as the RCNN threshold. Fourth, infer image boxes from trained models, save these boxes, and wait for the fusion of SwiftNet.
% There are 4 steps. First, generate instance boxes within images by 3D-2D projection and instance labels, and then save boxes to json file in the from of COCO dataset. Second, deploy MMdetection platform and load boxes json to register dataset. Third, feeding images of autopilot dataset,
% For the detector \textbf{result}, AP reachs 32.0\% for nuScenes while the one of SemanticKITTI is 30.8\%.

\noindent  \textbf{Module Cost} On nuScenes, the number parameters of baseline  is 296.2MB, and the memory cost is 26.2GB (batch size=2). With Cylinder-Mix, the memory and time cost (1.2 seconds per frame) remain nearly unchanged. When the IPSL module is added, memory and time costs acceptably increase by 3.5GB and 0.13 seconds per frame, respectively.
%------------------------------------------------------------------------

\subsection{Results of Detection Network}
To evaluate the Detection Network,we present the metrics between the ground-truth bounding boxes projected from 3D points and the predicted ones. As shown in Table \ref{table: results on detection network}, Faster R-CNN \cite{ren2015faster} with ResNet-50 and ResNet-101 \cite{he2016resnet} are the models for the detector. There are the average precision (AP) of each \textit{thing} class and mAP (\%) on the kittiImages and nuImages, which refer to the camera images of SemanticKITTI and nuScenes, respectively. The mAP of the Detection Network on both datasets appears to be low, this is due to the imbalance in the number of instances and the confusion of semantics. Nevertheless, its AP could reach 65.9\% if using a weighted sum by the number of instances. However, this also indicates the effectiveness of the IPSL module from another perspective. As these predicted instance boxes, as a form of weak labels, can still improve point cloud segmentation even without high precision.
% For example, motorcycle and motorcyclist similar in semantic classes, and their positions of ground truth bounding boxes are also very close. On SemanticKITTI, what’s more there are only 130 instances of motorcyclist, and its AP close to 0, which significantly lower the class average AP. Besides, despite of failing to distinguish semantic classes, it does not affect Instance information of our IPSL module, as the instance heatmap is only related to the predicted box position but not to semantics.
% trained detector

% \begin{table}[]
% \caption{Ablation experiments of IPSL and Cylinder-Mix on 100\% and 1\% SemanticKITTI}
% \label{Table: Ablation experiments on SemanticKITTI}
% \begin{tabular}{ccc|cc|ll}
% \toprule 
% \multicolumn{1}{l}{} & \multicolumn{1}{l}{} & \multicolumn{1}{l|}{} & \multicolumn{2}{c|}{SemanticKITTI} & \multicolumn{2}{c}{nuScenes}\\
% Baseline      & IPSL       & CM          & 40\%            & 1\%             & \multicolumn{1}{c}{40\%} & \multicolumn{1}{c}{1\%} \\ 
% \midrule
% $\surd$       &           &               & 60.4            & 46.5           & 75.5                     & 47.0                   \\
% $\surd$       & $\surd$   &               & 60.7            & 46.9           & 76.2                     & 48.5                   \\
% $\surd$       & $\surd$   & $\surd$       & 61.2            & 49.0           & 77.2                     & 49.8                   \\ 
% \bottomrule
% \end{tabular}
% \end{table}
\begin{center}
\begin{table}[]

\begin{tabular}{ccc|ccll}
\toprule 
Baseline      & IPSL      & CM          & 1\%    & 10\%   & 20\%   & 40\%  \\ 
\midrule
$\surd$       &           &             & 46.5   & 57.0   & 59.3   & 60.4   \\
$\surd$       & $\surd$   &             & 46.9   & 57.9   & 59.8   & 60.7  \\
$\surd$       & $\surd$   & $\surd$     & 49.0   & 59.9   & 60.8   & 61.2   \\ 
\bottomrule
\end{tabular}
\vspace{-3mm}
\caption{Ablation experiments of the IPSL module and Cylinder-Mix on SemanticKITTI}
\label{Table: Ablation experiments on SemanticKITTI}
\end{table}
\end{center}
%------------------------------------------------------------------------  
% \vspace{-8mm}
\begin{center}
\begin{table}[]

\begin{tabular}{c|l|cccc}
\toprule 
Scope                           &                 & 1\%   & 10\%  & 20\%  & 40\%  \\ \midrule
\multirow{2}{*}{total}          & baseline        & 46.5 & 57.0 & 59.3 & 60.4  \\
                                & baseline + IPSL & 46.9 & 57.9 & 59.8 & 60.7 \\ \hline
\multirow{2}{*}{inside}         & baseline        & 52.0 & 60.9 & 61.8 & 64.0 \\
                                & baseline + IPSL & 53.2 & 62.8 & 64.5 & 64.7 \\ \bottomrule
\end{tabular}
\vspace{-3mm}
\caption{Panoptic metrics on SemanticKITTI for total points and points inside the camera view.}
\label{table: eval_1_6}
% "SS-detector" is the Detector Network trained with semi-supervised data, while "fully detector" is trained with fully one. 
\end{table}
\end{center}
% \vspace{-5mm}
% ------------------------------------------------------------------------

\vspace{-12mm}
\subsection{Results of Segmentation on SemanctiKITTI}
As shown in table \ref{Table: Ablation experiments on SemanticKITTI}, we present the ablation results of the IPSL module and Cylinder-Mix components on SemanticKITTI.
We observe that Cylinder-Mix performs well on SemanticKITTI, particularly with improvements of 2.1\% and 2\% in PQ metric at 1\% and 10\% semi-supervised ratios, respectively. But the effect of the IPSL module is not as pronounced.
With a specific comparison on the 20\% SS ratio, Cylinder-Mix achieves a PQ gain of 2.1\% on nuScenes in the paper, while the gain on SemanticKITTI is only 0.5\%. It is probable that nuScenes has camera images with 6 views, which means that LiDAR-to-Camera projection and feature fusion of the IPSL module can cover most of the point cloud. In contrast, SemanticKITTI has only one front-facing camera, causing the IPSL module to affect a much smaller range of the point cloud, only 1/6 of the former.
% As shown in table \ref{Table: Ablation experiments on SemanticKITTI}, we present the ablation results of the IPSL module and Cylinder-Mix components on the SemanticKITTI, along with a comparison to the results on nuScenes on both 1\% and 40\% SS ratios. With specific comparison on the 1\% dataset, Cylinder-Mix achieves a PQ gain of 2.1\%, while the IPSL module is only 0.4\%, which corresponds to the nuScenes is 1.41\%. It is concluded that Cylinder Mix brings a significant performance promotion to SemanticKITTI, while the IPSL module does not show as much effectiveness as on nuScenes.  It is probable that nuScenes has camera images with 6 views, which results in that LiDAR-to-Camera projection and feature fusion of IPSL can cover most of the point cloud. In contrast, SemanticKITTI has only one front-facing camera, causing IPSL to affect a much smaller range of the point cloud, only 1/6 of the former.

Therefore, we introduce a new evaluation method for a fair comparison. To precisely assess the evaluation of IPSL on SemanticKITTI, we only calculate the PQ metrics for these points inside the camera view, cause IPSL primarily affects the points that can be projected onto the camera pixels. As shown in PQ metrics in Table \ref{table: eval_1_6}, through limited gain from the IPSL module for all points, it shows a noticeable improvement for points within the camera's field of view. This validates that IPSL primarily enhances the fusion of 2D-3D features and is actually effective.

\subsection{Ablation of IPSL Module}
\subsubsection{Instance Heatmap Encoder} Next, the parameters of the instance heatmap encoder have a direct impact on the heatmap, including $P_{center}$ that affects the size of the instance centre heatmap and the radius of the corner points $R_{corner}$. We conduct ablation experiments with different settings of the two parameters on the 1\% nuScenes. Also, we disable other data augmentation techniques to reduce model instability. As shown in Table \ref{table: ablation of boxes heatmap}, the experimental comparison is based on the performance of our baseline as setting 1, with a PQ of 47.0\%. Firstly, compared to setting 1 and the others, the IPSL module is effective under most parameters. Secondly, by comparing experiments 2, 3, and 4, we find that a too-small radius of the centre heatmap with $P_{center}$ is not conducive to indicating instance size information, and a parameter of $P_{center}=1/4$ is a good choice, resulting in a PQ of 48.5\%. Finally, by comparing experiments 2 and 5, we conclude that a too-large radius of the corner heatmap can decrease PQ, possibly because it causes confusion in indicating instance information.

%------------------------------------------------------------------------  
\begin{table}[]
\centering

\begin{tabular}{c|ccc}
\toprule 
setting     & $R_{corner}$ & $P_{center}$ &  PQ             \\ \midrule
1           & /         & /         & 47.0                \\ 
2           & 5         & 1/4       & 48.5               \\ 
% 3           & 5         & 1/4       & 48.22                \\ 
3           & 5         & 1/3       & 48.1                \\ 
4           & 5         & 1/10      & 47.9                \\ 
5           & 10        & 1/4       & 48.2                \\ \bottomrule
\end{tabular}
\vspace{-3mm}
\caption{Ablation study on boxes heatmap}
\label{table: ablation of boxes heatmap}
\vspace{-4mm}
\end{table}

% ------------------------------------------------------------------------
%-------------------------------------------------------------------------
\begin{table}[]
\begin{center}

\begin{tabular}{c|cc|c}
\toprule 
\multicolumn{1}{l|}{}        & $p_{cylmix}$ & $[R_x,R_y,R_z]$   &  PQ         \\ 
\midrule
baseline + IPSL              & 0        & /             & 46.5       \\ \hline
LaserMix                     & 0.25     & /             & 47.0       \\ \hline
\multirow{7}{*}{Cylinder-Mix}& 1        & {[}4,4,2{]}   & 47.5       \\
                             & 0.5      & {[}4,4,2{]}   & 47.4       \\
                             & 0.25     & {[}4,4,2{]}   & 49.0       \\
                             & 0.25     & {[}2,2,1{]}   & 46.9       \\
                             & 0.25     & {[}16,16,4{]} & 47.9       \\
                             & 0.25     & {[}1,1,4{]}   & 47.0       \\
                             & 0.25     & {[}8,8,1{]}   & 47.1       \\ 
\bottomrule
\end{tabular}
\vspace{-3mm}
\caption{Ablation of Cylinder-Mix on 1\% SemanticKITTI}
\label{table: ablation on Cylinder-Mix}
% \vspace{-5mm}
\end{center}
\end{table}
%-------------------------------------------------------------------------

\vspace{-2mm}
\subsection{Ablation of Cylinder-Mix}
As shown in the table \ref{table: ablation on Cylinder-Mix}, to explore the optimal parameters for using Cylinder-Mix(CM), we conducted ablation experiments on the mixing probability $p_{cylmix}$ and region size $[R_x,R_y,R_z]$ in SemanticKITTI. We choose 1\% as semi-supervised ratio, cause Cylinder-Mix is sensitive at this setting to distinguish effects in PQ under different parameters and the experimental comparison is based on the baseline with the IPSL module as a reference. \\ 
Firstly, to make a fair comparison with Lasermix, ring-mixing in Lasermix is performed as data augmentation on our baseline instead of Cylinder-Mix. At 25\%, LaserMix has a PQ of 47.0\%, while CM can achieve a maximum PQ of 49.0\%. Secondly, we fix the region size to [4,4,2] and adjusted the $p_{cylmix}$. The results demonstrate that $p_{cylmix}=1$ is not absolutely optimal and the model tends to benefit more from partially mixed data with original data rather than entirely mixed data. Finally, we fixed the $p_{cylmix}$ to 25\% and adjusted the region size to [4,4,2], [16,16,4], [1,1,4] and [8,8,1], respectively. It is expected to evaluate the sensitivity of Cylinder-Mix along with the Z-axis and X-Y axes. We don't separately increase the region size of the X and Y axes because they are equivalent in the cylinder voxel. Based on the experimental results, we draw the conclusion that adding region size solely at $R_z$  or the XY-plane is not the optimal mixing strategy, and too few regions to mix are also not ideal. A moderately large region size, such as [4,4,2], is a good choice of region size.

\section{Supplementary Experiments}
% \subsection{Additional mIoU benefited from instance labels}

% Additionally, it may be unfair to compare the semantic segmentation model with our model which uses instance labels. Therefore, to estimate the additional mIoU benefits from instance labels, we train semantic segmentation on our baseline only using semantic labels and then compare the mIoU with that of Panoptic baseline. In detail, We remove the instance-related predictions in the multi-Heads and only utilize the semantic head for training.

% As shown in Table \ref{table: mIoU gain}, we present the comparison of mIoU results on fully supervised and 1\% semi-supervised settings on both nuScenes and SemanticKITTI. Semantic baseline refers to the model trained for semantic segmentation using only semantic labels, while Panoptic baseline refers to the model trained for panoptic segmentation using both semantic and instance labels. From the table, we can see that on the nuScenes dataset, there is a mIoU gain of approximately 1.6\textasciitilde1.7\% attributed to the instance labels, while on the SemanticKITTI dataset, this mIoU gain ranges from 0.8\textasciitilde1.2\%. Building on this conclusion, our model's performance still surpasses LaserMix \cite{kong2023lasermix} even when accounting for the additional mIoU benefited from instance labels.

\subsection{Improvement of RGB images}
For RGB information, we removed the image branch of our baseline to observe its contribution on the fully supervised datasets. As shown in Table \ref{table: RGB information}, the RGB information results in a PQ gain of 2.1\% and an mIoU gain of 2.9\% on nuScenes. However, the improvement is not significant on the SemanticKITTI, it may be attributed to the fact that SemanticKITTI has only 1/6th of the angle of camera view compared to nuScenes.

%------------------------------------------------------------------------  
\begin{table}[]

\begin{center}
\begin{tabular}{c|l|cc}
\toprule 
                      & \multicolumn{1}{c|}{}     & nuScenes & SemanticKITTI \\ \midrule
\multirow{2}{*}{PQ}   & w/o IB & 75.4    & 59.1         \\
                      & baseline                  & 77.6    & 59.8         \\ \hline
\multirow{2}{*}{mIoU} & w/o IB & 77.9    & 63.7         \\
                      & baseline                 & 80.7    & 64.2         \\ \bottomrule
\end{tabular}
\vspace{-3mm}
\caption{Ablation study of RGB information on fully supervised dataset. w/o IB: baseline without Image Branch.}
\label{table: RGB information}
\end{center}
\vspace{-3mm}
\end{table}
% ------------------------------------------------------------------------

%------------------------------------------------------------------------  
% \begin{table}[]
% \begin{center}
% \vspace{-3mm}
    
% \caption{mIoUs of panoptic and semantic baseline.}
% \label{table: mIoU gain}
% \begin{tabular}{c|cc|cc}
% \toprule 
%                   & \multicolumn{2}{c|}{nuScenes} & \multicolumn{2}{c}{SemanticKITTI} \\
%                   & 100\%         & 1\%           & 100\%           & 1\%             \\ \midrule
% Semantic baseline & 80.5         & 54.5          & 63.0              & 50.3           \\
% Panoptic baseline & 82.1         & 56.2         & 64.2           & 51.2           \\ 
% \bottomrule
% \end{tabular}
% \end{center}
% \end{table}
% ------------------------------------------------------------------------

%-----------------------------------------------------------------------
% \begin{figure}
% % \fbox{\rule{0pt}{2in} \rule{.9\linewidth}{0pt}}
% \includegraphics[width=\linewidth]{samples/appendix/epoch_PQs.pdf}
% \caption{The evolution of PQ over epoch} 
%  % for different settings on SemanticKITTI
% \label{epoch_PQs}
%    % \caption{The framework of .}
% \end{figure}

%------------------------------------------------------------------------
\subsection{Details PQ and IoU scores}
In Table \ref{table: Class-wise IoU and PQ}, we present the class-wise IoU and PQ on the val set of the SemanticKITTI under the 100\% and 10\% settings. The table indicates that our model does not show significant improvements in the \textit{stuff} category, but shows a clear increase in PQ for the \textit{thing} category, which indicates that the proposed IPSL module and Cylinder-Mix put more emphasis on segmentation of \textit{thing} classes.

% ------------------------------------------------------------------------
\section{Visualization}
\subsection{SAM mask}
\begin{figure}[]
\begin{center}
\includegraphics[width=1\linewidth]{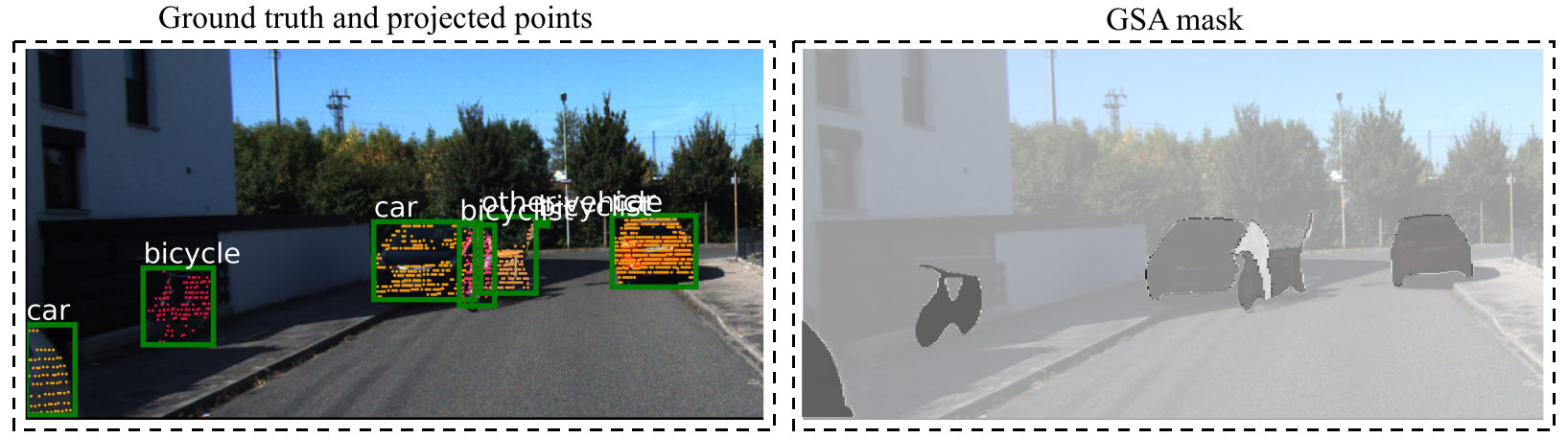}
\end{center}
   \vspace{-5mm}
   \caption{Ground truth and high-quality GSA mask.}
\label{GSA mask}
\vspace{-3mm}
\end{figure}

For each frame of the camera image, we enter each category name of the dataset as prompt. With SAM masks (Grounded Segment Anything, GSA shown in Figure \ref{GSA mask}), our approach achieves very close performance 70.7@mIoU compared with the detector 70.9@mIoU in Table 4 of original paper. That demonstrates IPSL is able to combine with vision foundation model.

% \subsection{evolution of PQ over epoch}
% Based on the \textit{Implementation Details} of Point Cloud Panoptic Segmentation, the number of epochs during \textbf{pretrain} increases as the semi-supervised ratio decreases. As shown is Figure \ref{epoch_PQs}, We present the variation of PQs with epochs on the SemanticKITTI dataset and indicate the position where the maximum PQ is achieved on the validation set. The results show that the number of epochs designed2 is reasonable. When the amount of labeled data decreases, the model needs to increase the epoch appropriately to increase the number of gradient descent updates for sufficient training.

\subsection{segmentation visualization of point clouds}
\begin{figure*}
\begin{center}
% \fbox{\rule{0pt}{2in} \rule{.9\linewidth}{0pt}}
\includegraphics[width=1\linewidth]{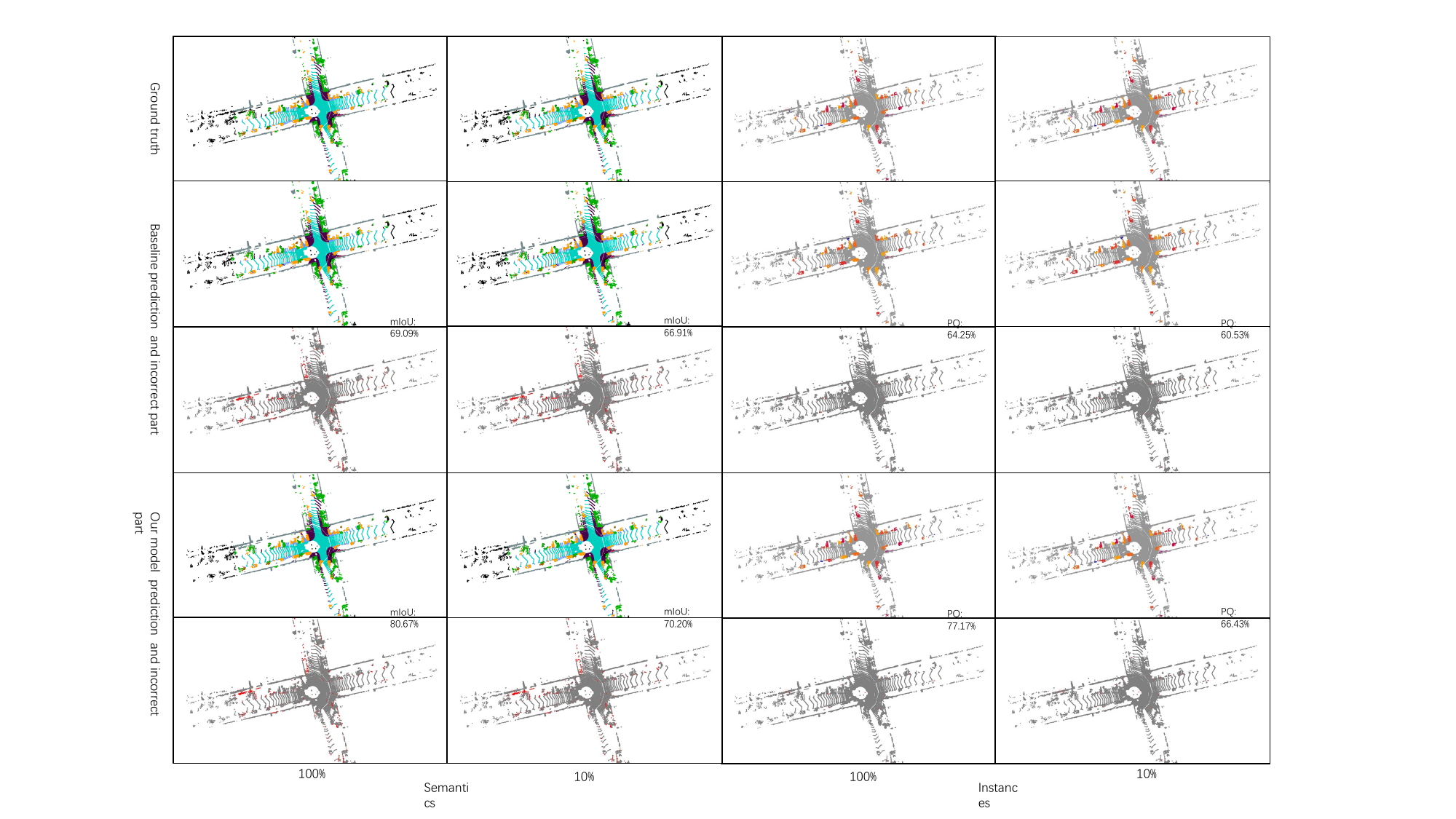}
\caption{Visualization on SemanticKITTI}
\label{Visualization on SemanticKITTI}
\end{center}
   % \caption{The framework of .}
   
\end{figure*}

As shown in Figure \ref{Visualization on SemanticKITTI}, we provide visualizations of point clouds for the ground truth, predictions of the baseline, and our model. The color red is used to indicate incorrect predictions in the point cloud. Semantic segmentation and instance segmentation visualisations are presented. The mIoU or PQ excluding categories that are absent within ground truth is labeled in the bottom-right corner of the subplot. The 100\% on the left indicates the model trained on the full SemanticKITTI dataset, while the 10\% on the right means only using on 10\% labeled subset following our SS setting.

\begin{table*}[]

\begin{center}
% \small
% \footnotesize
\scriptsize
\begin{tabular}{c|cccccccccc|c|c}
\toprule 
kittiImages           & car   & bicycle & truck & person                                                 & \begin{tabular}[c]{@{}c@{}}motor-\\ cyclist\end{tabular} & bicyclist & \begin{tabular}[c]{@{}c@{}}motor-\\ cyclle\end{tabular} & \begin{tabular}[c]{@{}c@{}}other-\\ vehicle\end{tabular}        &         &                                                        & \begin{tabular}[c]{@{}c@{}}sum or\\ mAP\end{tabular} & \begin{tabular}[c]{@{}c@{}}weighted \\ mAP\end{tabular} \\ 
\midrule
Number of instances   & 36115 & 2602    & 258   & 3334                                                   & 130                                                      & 1488      & 897                                                     & 2230                                                            &         &                                                        & 43927                                                &                                                         \\ \hline
Faster R-CNN + ResNet-50  & 69.7  & 16.4    & 14.4  & 28.2                                                   & 0.1                                                      & 52.8      & 17.7                                                    & 21.4                                                            &         &                                                        & 27.5                                                 & 63.7                                                    \\
Faster R-CNN + ResNet-101 & 71.4  & 21      & 20.6  & 27.8                                                   & 0                                                        & 56.8      & 21.3                                                    & 27.5                                                            &         &                                                        & 30.8                                                 & 65.9                                                    \\ 
\midrule
nuImages              & car   & bicycle & truck & \begin{tabular}[c]{@{}c@{}}pedes-\\ trian\end{tabular} & \begin{tabular}[c]{@{}c@{}}motor-\\ cycle\end{tabular}   & bus       & barrier                                                 & \begin{tabular}[c]{@{}c@{}}construction \\ vehicle\end{tabular} & trailer & \begin{tabular}[c]{@{}c@{}}traffic\\ cone\end{tabular} & \begin{tabular}[c]{@{}c@{}}sum or\\ mAP\end{tabular} & \begin{tabular}[c]{@{}c@{}}weighted \\ mAP\end{tabular} \\ 
\midrule
Number of instances   & 70398 & 2201    & 16407 & 33180                                                  & 2440                                                     & 3449      & 25059                                                   & 2716                                                            & 4451    & 12892                                                  & 160301                                               &                                                         \\ \hline
Faster R-CNN + ResNet-50  & 44.6  & 21.7    & 39.2  & 25.5                                                   & 26.9                                                     & 54.3      & 33.6                                                    & 15.2                                                            & 27.8    & 17.5                                                   & 30.6                                                 & 38.4                                                    \\
Faster R-CNN + ResNet-101 & 45    & 21.3    & 40.3  & 25.6                                                   & 27.6                                                     & 55.4      & 34.3                                                    & 16.7                                                            & 28.5    & 17.4                                                   & 31.2                                                 & 38.9                                                    \\ 

\bottomrule
\end{tabular}
\end{center}
\vspace{-3mm}
\caption{Results on Detection Network of different backbone}
\label{table: results on detection network}

\end{table*}
% %------------------------------------------------------------------------

%------------------------------------------------------------------------  
\begin{table*}[]
\begin{center}
\tiny
\resizebox{\textwidth}{15mm}{
\begin{tabular}{c|cc|cccccccccccccccccccccc}
\toprule 
                       &                           & metrics & car   & bicy  & moto  & truck & bus   & bus   & ped   & b.cyc & PQ\_th & road  & park  & walk  & o.gro & build & fence & veg   & trunk  & terr  & pole  & sign  & PQ\_st & mean \\ \midrule
\multirow{4}{*}{100\%} & \multirow{2}{*}{baseline} & IoU     & 94.1  & 54.71 & 72.55 & 75.01 & 45.98 & 72.31 & 84.78 & 29.09 & 66.07  & 94.32 & 44.59 & 80.82 & 3.02  & 88.49 & 45.79 & 85.83 & 64.57  & 69.58 & 65.5  & 47.81 & 62.76  & 64.15\\
                       &                           & PQ      & 90.29 & 58.05 & 70.63 & 57.79 & 48.44 & 80.02 & 87.82 & 35.05 & 66.01  & 94.26 & 23.71 & 77.48 & 0.13  & 86.92 & 16.75 & 83.54 & 49.3   & 55.79 & 61.57 & 58.16 & 55.238 & 59.77\\ \cline{2-25} 
                       & \multirow{2}{*}{ourmodel} & IoU     & 94.37 & 57.46 & 73.93 & 76.36 & 59.11 & 76.37 & 88.17 & 26.07 & 68.98  & 93.5  & 39.65 & 80.33 & 2     & 89.44 & 51.21 & 85.79 & 65.12  & 69.73 & 65.4  & 47.27 & 62.68  & 65.33\\
                       &                           & PQ      & 90.38 & 60.53 & 74.80 & 62.10 & 58.33 & 82.85 & 89.49 & 31.36 & 68.73  & 93.74 & 22.10 & 77.46 & 0     & 87.04 & 19.95 & 84.23 & 49.62  & 54.23 & 60.92 & 56.18 & 55.04  & 60.81\\ \hline
\multirow{4}{*}{10\%}  & \multirow{2}{*}{baseline} & IoU     & 92.5  & 50.96 & 60.66 & 73.77 & 39.38 & 69.01 & 78.74 & 12.76 & 59.72  & 94.15 & 43.46 & 80.79 & 2.59  & 89.71 & 52.26 & 86.54 & 64.85  & 70.49 & 66.24 & 45.47 & 63.32  & 61.81\\
                       &                           & PQ      & 88.26 & 52.36 & 64.19 & 46.19 & 42.19 & 76.25 & 87.75 & 19.71 & 59.61  & 93.97 & 22.27 & 76.81 & 0.13  & 86.87 & 19.28 & 84.93 & 49.63  & 54.84 & 62.17 & 54.72 & 55.06  & 56.98\\ \cline{2-25} 
                       & \multirow{2}{*}{ourmodel} & IoU     & 94.21 & 54.66 & 68.19 & 76.08 & 49.83 & 73.28 & 83.19 & 26.19 & 65.70  & 94.64 & 43.43 & 81.43 & 0.41  & 90.75 & 52.17 & 87.58 & 66.67  & 72.63 & 66.48 & 48.44 & 64.06  & 64.75\\
                       &                           & PQ      & 89.68 & 56.15 & 71.20 & 55.63 & 50.19 & 80.82 & 87.86 & 27.96 & 64.94  & 94.56 & 25.46 & 77.96 & 0.46  & 87.93 & 19.16 & 85.32 & 52.16  & 55.37 & 61.68 & 57.80 & 56.17  & 59.86\\ 
\bottomrule
\end{tabular}}
\end{center}
\vspace{-3mm}
\caption{Class-wise IoU and PQ scores on the val set of SemanticKITTI.}
\label{table: Class-wise IoU and PQ}
\end{table*}
% ------------------------------------------------------------------------

%------------------------------------------------------------------------

%------------------------------------------------------------------------

\end{document}